%% file: iccv_draft.tex
\documentclass[10pt,twocolumn,letterpaper]{article}

\usepackage{iccv}
\usepackage{times}
\usepackage{epsfig}
\usepackage{graphicx}
\usepackage{amsmath}
\usepackage{amssymb}
\usepackage{makecell}
\usepackage{caption}
\input{math_commands.tex}

\usepackage{marvosym}

\usepackage{bbm}
\usepackage{amsfonts}
\usepackage{amsthm}
\usepackage{multirow}
\usepackage{enumitem}
\usepackage{framed}
\usepackage{mathtools} 
\usepackage{breqn}
\usepackage{footmisc}
\usepackage{makecell}
\usepackage{footnote}
\makesavenoteenv{tabular}
\makesavenoteenv{table}
\usepackage[pagebackref=true,breaklinks=true,letterpaper=true,colorlinks,bookmarks=false]{hyperref}

\iccvfinalcopy % *** Uncomment this line for the final submission

\def\iccvPaperID{1152} % *** Enter the ICCV Paper ID here

\begin{document}

%%%%%%%%% TITLE
\title{Talk-to-Edit: Fine-Grained Facial Editing via Dialog}

\author{Yuming Jiang$^{1*}$
\quad
Ziqi Huang$^{1*}$
\quad
Xingang Pan$^{2}$
\quad
Chen Change Loy$^{1}$
\quad
Ziwei Liu\textsuperscript{1\Letter}\\
$^{1}$S-Lab, Nanyang Technological University
\quad
$^{2}$The Chinese University of Hong Kong
\\
{\tt\small \{yuming002, hu0007qi, ccloy, ziwei.liu\}@ntu.edu.sg \hspace{5pt} px117@ie.cuhk.edu.hk}\\
}

% \maketitle

\twocolumn[{%
            \renewcommand\twocolumn[1][]{#1}%
            \vspace{-1em}
            \maketitle
            \vspace{-3em}
            \begin{center}
                \centering
                % \vspace{-0.3in}
                \includegraphics[width=0.99\textwidth]{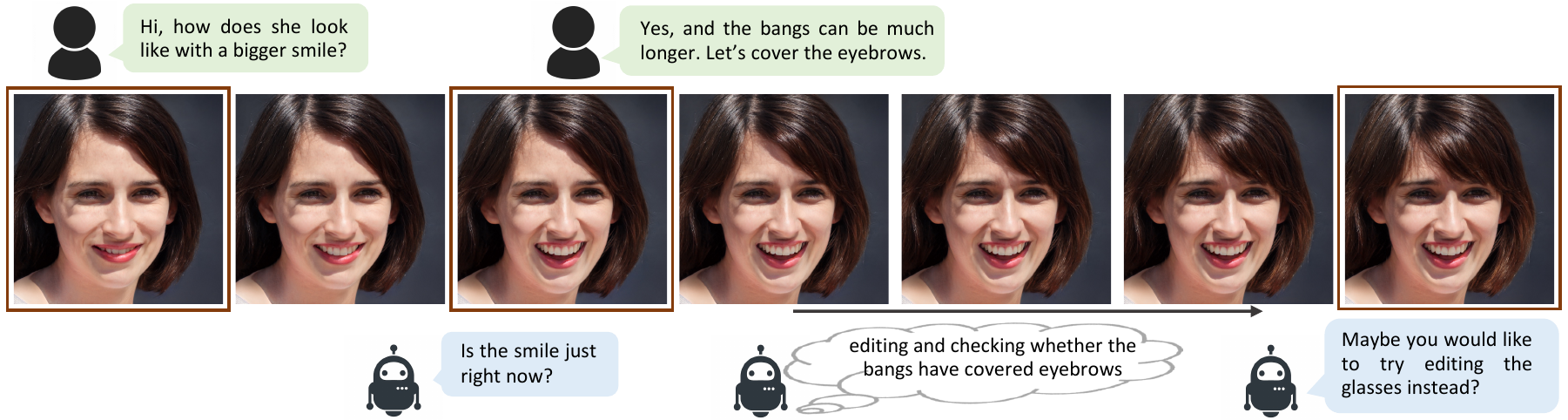}\vspace{0.1cm}
                \vskip -0.15cm
                \captionof{figure}{\textbf{An example of \textit{Talk-to-Edit}}. The user provides a facial image and an editing request. Our system then edits the image accordingly, and provides meaningful language feedback such as clarification or alternative editing suggestions. During editing, the system is able to control the extent of attribute change on a fine-grained scale, and iteratively checks whether the current editing step fulfills the user's request.}
                \label{teaser}
            \end{center}%
        }]

% \thispagestyle{empty} 

%%%%%%%%% ABSTRACT
\begin{abstract}

   Facial editing is an important task in vision and graphics with numerous applications. However, existing works are incapable to deliver a continuous and fine-grained editing mode (e.g., editing a slightly smiling face to a big laughing one) with natural interactions with users.
   In this work, we propose \textbf{Talk-to-Edit}, an interactive facial editing framework that performs fine-grained attribute manipulation through dialog between the user and the system.
   Our key insight is to model a continual ``semantic field'' in the GAN latent space.
   \textbf{1)} Unlike previous works that regard the editing as traversing straight lines in the latent space, here the fine-grained editing is formulated as finding a curving trajectory that respects fine-grained attribute landscape on the semantic field. \textbf{2)} The curvature at each step is location-specific and determined by the input image as well as the users' language requests. \textbf{3)} To engage the users in a meaningful dialog, our system generates language feedback by considering both the user request and the current state of the semantic field.
  \makeatletter{\renewcommand*{\@makefnmark}{}
  \footnotetext{$^*$Equal contribution.}\makeatother}
   
   We also contribute \textbf{CelebA-Dialog}, a visual-language facial editing dataset to facilitate large-scale study. 
   Specifically, each image has manually annotated fine-grained attribute annotations as well as template-based textual descriptions in natural language.
   Extensive quantitative and qualitative experiments demonstrate the superiority of our framework in terms of \textbf{1)} the smoothness of fine-grained editing, \textbf{2)} the identity/attribute preservation, and \textbf{3)} the visual photorealism and dialog fluency. Notably, user study validates that our overall system is consistently favored by around $80\%$ of the participants. Our project page is \url{https://www.mmlab-ntu.com/project/talkedit/}.    
   
\end{abstract}
\vspace{-0.5cm}

\input{section/introduction}

\input{section/related_work}
\input{section/dataset}
\input{section/approach}
\input{section/experiment}
\input{section/conclusion}

{\small
\bibliographystyle{ieee_fullname}
\bibliography{egbib}
}
\newpage
\input{section/iccv_supp}

\end{document}

%% file: math_commands.tex
\usepackage{amsmath,amsfonts,bm}

\def\eqref#1{equation~\ref{#1}}
\def\1{\bm{1}}

\def\rmI{{\mathbf{I}}}

\def\vzero{{\bm{0}}}

\def\ve{{\bm{e}}}
\def\vf{{\bm{f}}}

\def\vh{{\bm{h}}}

\def\vr{{\bm{r}}}
\def\vs{{\bm{s}}}

\def\vz{{\bm{z}}}

\DeclareMathAlphabet{\mathsfit}{\encodingdefault}{\sfdefault}{m}{sl}
\SetMathAlphabet{\mathsfit}{bold}{\encodingdefault}{\sfdefault}{bx}{n}

%% file: section/introduction.tex
\section{Introduction}

The goal of facial editing is to enable users to manipulate facial images in their desired ways.
Thanks to the advance of deep generative models like GANs \cite{gan, conditionalgan, biggan, pggan, stylegan1, stylegan2}, facial editing has witnessed rapid growth in recent years, especially in image fidelity.
While there have been several attempts to improve facial editing quality, they often lack interactions with users or require users to follow some fixed control patterns.
For instance, image-to-image translation models \cite{cyclegan, stargan, translation_conditional, cross_domain, unsupervised_translation} only translate facial images between several discrete and fixed states, and users cannot give any subjective controls to the system.
Other face editing methods offer users some controls, such as a semantic map indicating the image layout \cite{maskgan}, a reference image demonstrating the target style \cite{perceptual, universal_styletransfer, yin2019instance, li2021image}, and a sentence describing a desired effect \cite{recurrent_attentive, text_as_neural_operator, text_adaptive, prada, tedigan}.
However, users have to follow the fixed patterns, which are too demanding and inflexible for most users.
Besides, the only feedback provided by the system is the edited image itself.

In terms of the flexibility of interactions, we believe natural language is a good choice for users. 
Language is not only easy to express and rich in information, but also a natural form for the system to give feedback.
Thus, in this work, we make the first attempt towards a dialog-based facial editing framework, namely \textbf{Talk-to-Edit}, where editing is performed round by round via request from the user and feedback from the system.

In such an interactive scenario, users might not have a clear target in their mind at the beginning of editing and thoughts might change during editing, like tuning an overly laughing face back to a moderate smile.
Thus, the editing system is supposed to be capable of performing continuous and fine-grained attribute manipulations. 
While some approaches \cite{interfacegan1, interfacegan2, unsupervised_discovery, closed_form_factorization, pca} could perform continuous editing to some extent by shifting the latent code of a pre-trained GAN \cite{stylegan1, stylegan2, pggan, biggan}, they typically make two assumptions: 1) the attribute change is achieved by traversing along a straight line in the latent space; 2) different identities share the same latent directions.
However, these assumptions overlook the non-linear nature of the latent space of GAN, potentially leading to several shortcomings in practice: \textbf{1)} The identity would drift during editing; \textbf{2)} When editing an attribute of interest, other irrelevant attributes would be changed as well; \textbf{3)} Artifacts would appear if the latent code goes along the straight line too far. 

To address these challenges, we propose to learn a \textit{vector field} that describes \textit{location-specific} directions and magnitudes for attribute changes in the latent space of GAN, which we term as a ``semantic field''.
Traversing along the curved trajectory takes into account the non-linearity of attribute transition in the latent space, thus achieving more fine-grained and accurate facial editing.
Besides, the curves changing the attributes of different identities might be different, which can also be captured by our semantic field with the location-specific property.
In this case, the identity of the edited facial image would be better preserved. 
In practice, the semantic field is implemented as a mapping network, and is trained with fine-grained labels to better leverage its location-specific property, which is more expressive than prior methods supervised by binary labels.

The above semantic field editing strategy is readily embedded into our dialog system to constitute the whole \textit{Talk-to-Edit} framework.
Specifically, a user's language request is encoded by a language encoder to guide the semantic field editing part to alter the facial attributes consistent with the language request.
After editing, feedback would be given by the system conditioned on previous edits to check for further refinements or offer other editing suggestions.
The user may respond to the system feedback for further editing actions, and this dialog-based editing iteration would continue until the user is satisfied with the edited results.

To facilitate the learning of semantic field and dialog-based editing, we contribute a large-scale visual-language dataset named \textbf{CelebA-Dialog}. 
Unlike prior datasets with only binary attribute labels, we annotate images in CelebA with attribute labels of fine granularity.
Accompanied with each image, there is also a user request sample and several captions describing these fine-grained facial attributes.

In summary, our main contributions are: 
\textbf{1)} We propose to perform fine-grained facial editing via dialog, an easier interactive way for users.
\textbf{2)} To achieve more continuous and fine-grained facial editing, we propose to model a location-specific semantic field.
\textbf{3)} We achieve superior results with better identity preservation and smoother change compared to other counterparts.
\textbf{4)} We contribute a large-scale visual-language dataset \textbf{CelebA-Dialog}, containing fine-grained attribute labels and textual descriptions.

%% file: section/related_work.tex
\begin{figure*}
   \begin{center}
      \includegraphics[width=1.0\linewidth]{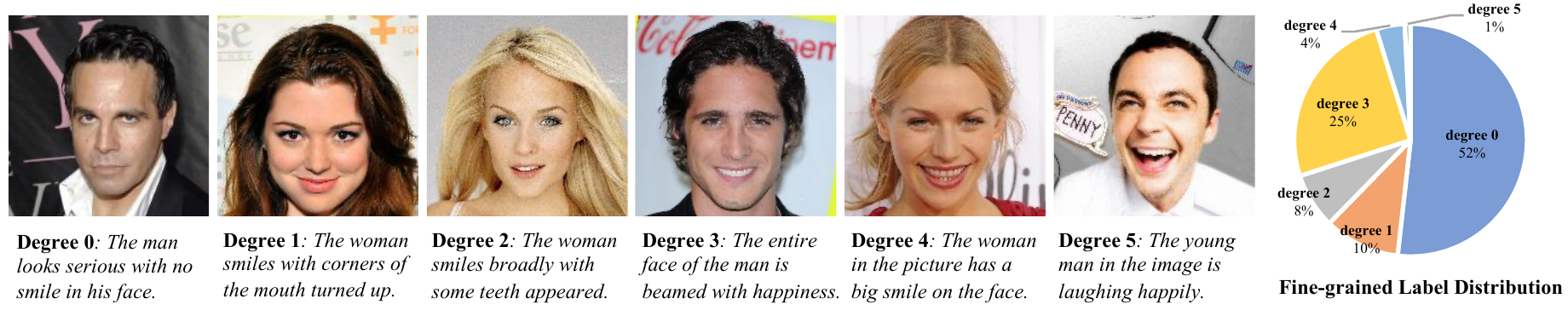}
   \end{center}
   \vspace{-14pt}
   \caption{\textbf{Illustration of \textit{CelebA-Dialog} dataset.} 
   We show example images and annotations for the smiling attribute. 
   Below the images are the attribute degrees and the corresponding textual descriptions. We also show the fine-grained label distribution of the smiling attribute.}
   \label{dataset_illustration}
   \vspace{-14pt}
\end{figure*}

\section{Related Work}

%%%%%%%%%%%%%%%%%%%%%%%%%%%%%%%%%%%%%%%%%%%%%%%%%%%%%%%%%
\noindent\textbf{Semantic Facial Editing.} 
Several methods have been proposed for editing specific attributes such as age progression \cite{age_progression_gan, recurrent_face_aging}, hair synthesis \cite{hair_synthesis, hair_modeling}, and smile generation \cite{smile_generation}. 
Unlike these attribute-specific methods relying on facial priors such as landmarks, our method is able to manipulate multiple semantic attributes without using facial priors.
Image-to-image translation methods \cite{cyclegan, stargan, translation_conditional, cross_domain, unsupervised_translation} have shown impressive results on facial editing.
However, they are insufficient to perform continuous editing because images are translated between two discrete domains. 

%%%%%%%%%%%%%%%%%%%%%%%%%%%%%%%%%%%%%%%%%%%%%%%%%%%%%%%%%
Recently, latent space based manipulation methods \cite{zhu2016generative, brock2016neural} are drawing increasing attention due to the advancement of GAN models like StyleGAN \cite{karras2019style,karras2020analyzing}.
These approaches typically discover semantically meaningful directions in the latent space of a pretrained GAN so that moving the latent code along these directions could achieve desired editing in the image space. Supervised methods find directions to edit the attributes of interest using attribute labels \cite{interfacegan1, interfacegan2, enjoy_your_editing}, while unsupervised methods exploit semantics learned by the pretrained GAN to discover the most important and distinguishable directions \cite{unsupervised_discovery, pca, closed_form_factorization}.
InterFaceGAN \cite{interfacegan1, interfacegan2} finds a hyperplane in the latent space to separate semantics into a binary state and then uses the normal vector of the hyperplane as the editing direction.
A recent work \cite{enjoy_your_editing} learns a transformation supervised by binary attribute labels and directly adds the transformation direction to the latent code to achieve one-step editing. Some approaches \cite{jahanian2019steerability, abdal2021styleflow} consider the non-linear property of latent space.
Different from existing methods, we learn a location-specific field in the latent space supervised by fine-grained labels to achieve precise fine-grained editing and to preserve facial identities.

%%%%%%%%%%%%%%%%%%%%%%%%%%%%%%%%%%%%%%%%%%%%%%%%%%%%%%%%%
\noindent\textbf{Language-based Image Editing.}
The flexibility of natural language has attracted researchers to propose a number of text-to-image generation \cite{stackgan, generative_text_to_image, attngan, tedigan} and manipulation \cite{recurrent_attentive, text_as_neural_operator, text_adaptive, prada, tedigan} approaches.
For example, given an input image, TediGAN \cite{tedigan} generates a new image conditioned on a text description.
Some other approaches \cite{benchmark, sequential_attngan, codraw, neural_painter, sscr, liu2020describe, li2020manigan} allow users to give requests in the form of natural language but do not provide meaningful feedback, clarification, suggestion, or interaction. Chatpainter \cite{chatpainter} synthesizes an image conditioned on a completed dialog, but could not talk to users round by round to edit images. 
Unlike existing systems that simply ``listen" to users to edit, our dialog-based editing system is able to ``talk" to users, edit the image according to user requests, clarify with users about their intention especially on fine-grained attribute details, and offer other editing options for users to explore.

%% file: section/dataset.tex
\section{CelebA-Dialog Dataset}

In the dialog-based facial editing scenarios, many rounds of edits are needed till users are satisfied with the edited images. To this end, the editing system should be able to generate continuous and fine-grained facial editing results, which contain intermediate states translating source images to target images. 
However, for most facial attributes, binary labels are not enough to precisely express the attribute degrees.
Consequently, methods trained with only binary labels could not perform natural fine-grained facial editing. 
Specifically, they are not able to generate plausible results when attribute degrees become larger.
Thus, fine-grained facial attribute labels are vital to providing supervision for fine-grained facial editing.
Moreover, the system should also be aware of the attribute degrees of edited images so that it could provide precise feedback or suggestions to users, which also needs fine-grained labels for training. 

Motivated by these, we contribute a large-scale visual-language face dataset named \textbf{CelebA-Dialog}. 
The \textbf{CelebA-Dialog} dataset has the following properties: 
\textbf{1)} Facial images are annotated with rich fine-grained labels, which classify one attribute into multiple degrees according to its semantic meaning; 
\textbf{2)} Accompanied with each image, there are captions describing the attributes and a user request sample. 
The \textbf{CelebA-Dialog} dataset is built as follows:

\begin{figure*}
   \begin{center}
      \includegraphics[width=1.0\linewidth]{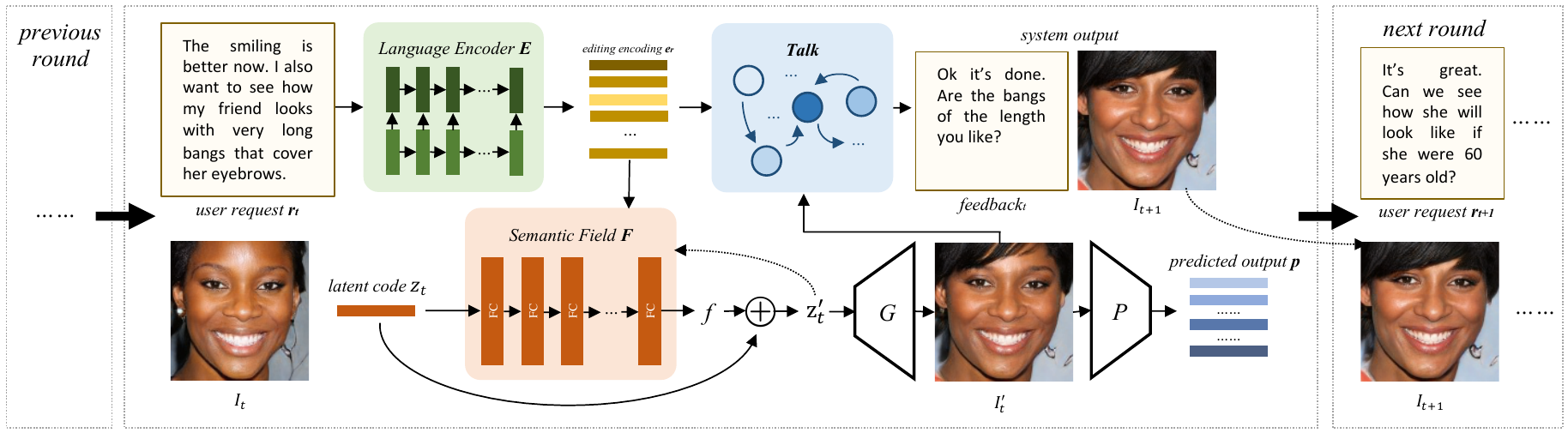}
   \end{center}
   \vspace{-16pt}
   \caption{\textbf{Overview of \textit{Talk-to-Edit} Pipeline.} 
   In round $t$, we receive the input image $\rmI_t$ and its corresponding latent code $\vz_t$ from the last round.
   Then the \textit{Language Encoder} $E$ extracts the editing encoding $\ve_r$ from the user request $\vr_t$, and feeds $\ve_r$ to the \textit{Semantic Field} $F$ to guide the editing process. 
   The latent code $\vz_t$ is iteratively moved along field lines by adding the field vector $\vf=F(\vz_t)$ to $\vz_t$, and a pretrained predictor is used to check whether the target degree is achieved. 
   Finally, the edited image $\rmI_{t+1}$ will be output at the end of one round. 
   Based on the editing encoding $\ve_r$, the $Talk$ module gives language feedback such as clarification and alternative editing suggestions.}
   \label{pipeline}
   \vspace{-9pt}
\end{figure*}

\noindent\textbf{Data Source.} 
CelebA dataset \cite{celeba} is a well-known large-scale face attributes dataset, which contains 202,599 images. With each image, there are forty binary attribute annotations. Due to its large-scale property and diversity, we choose to annotate fine-grained labels for images in CelebA dataset. Among forty binary attributes, we select five attributes whose degrees cannot be exhaustively expressed by binary labels. The selected five attributes are Bangs, Eyeglasses, Beard, Smiling, and Young (Age). 

\noindent\textbf{Fine-grained Annotations.}
For Bangs, we classify the degrees according to the proportion of the exposed forehead. There are 6 fine-grained labels in total: 100\%, 80\%, 60\%, 40\%, 20\%, and 0\%. 
The fine-grained labels for eyeglasses are annotated according to the thickness of glasses frames and the type of glasses (ordinary / sunglasses). 
The annotations of beard are labeled according to the thickness of the beard. And the metrics for smiling are the ratio of exposed teeth and open mouth.
As for the age, we roughly classify the age into six categories: below 15, 15-30, 30-40, 40-50, 50-60, and above 60.
In Fig.~\ref{dataset_illustration}, we provide examples on the fine-grained annotations of the smiling attribute.
For more detailed definitions and examples of fine-grained labels for each attribute, please refer to the supplementary files.

\noindent\textbf{Textual Descriptions.}
For every image, we provide fine-grained textual descriptions which are generated via a pool of templates. The captions for each image contain one caption describing all the five attributes and five individual captions for each attribute. 
Some caption examples are given in Fig.~\ref{dataset_illustration}.
Besides, for every image, we also provide an editing request sample conditioned on the captions. For example, a serious-looking face is likely to be requested to add a smile.

%% file: section/approach.tex
\section{Our Approach}

The pipeline of \textbf{Talk-to-Edit} system is depicted in Fig. \ref{pipeline}. The whole system consists of three major parts: user request understanding, semantic field manipulation, and system feedback. The initial inputs to the whole system are an image $\rmI$ and a user's language request $\vr$. A language encoder $E$ is first employed to interpret the user request into the editing encoding $\ve_{r}$, indicating the attribute of interest, changing directions, etc. Then the editing encoding $\ve_{r}$ and the corresponding latent code $\vz$ is fed into the ``semantic field'' $F$ to find the corresponding vectors $\vf_{z}$ to change the specific attribute degrees. After one round of editing, the system will return the edited image $\rmI^{\prime}$ and provide reasonable feedback to the user. 
The editing will continue until the user is satisfied with the editing result.

\subsection{User Request Understanding}

Given a user's language request $\vr$, we use a language encoder $E$ to extract the editing encoding $\ve_{r}$ as follows:
\begin{equation}
    \ve_{r} = E(\vr)
\end{equation}
The editing encoding $\ve_{r}$, together with the dialog and editing history, and the current state of the semantic field, will decide and instruct the semantic field whether to perform an edit in the current round of dialog. The editing encoding $\ve_{r}$ contains the following information:
\textbf{1)} request type,
\textbf{2)} the attribute of interest, 
\textbf{3)} the editing direction,
and \textbf{4)} the change of degree.

Users' editing requests  are classified into three types: 
\textbf{1)} describe the attribute and specify the target degree, 
\textbf{2)} describe the attribute of interest and indicate the relative degree of change, 
\textbf{3)} describe the attribute and only the editing direction without specifying the degree of change.
We use template-based method to generate the three types of user requests and then train the language encoder.

\subsection{Semantic Field for Facial Editing}

\begin{figure*}
   \begin{center}
      \includegraphics[width=0.9\linewidth]{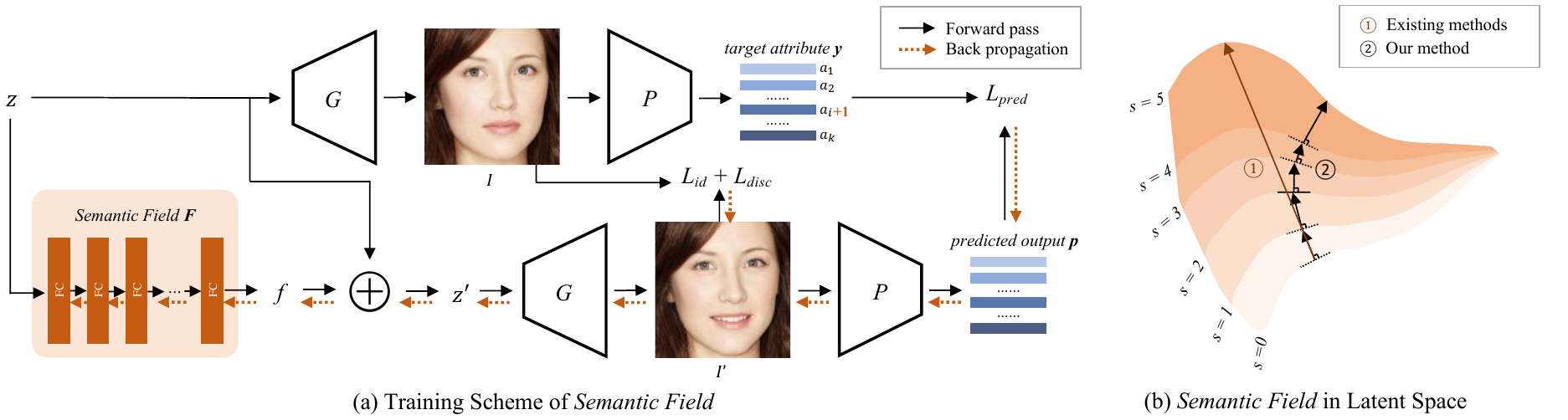}
   \end{center}
   \vspace{-0.6cm}
   \caption{\textbf{(a) Training Scheme of \textit{Semantic Field}. } Predictor loss, identity keeping loss and discriminator loss are adopted to ensure the location-specific property of semantic field. \textbf{(b) Illustration of \textit{Semantic Field} in Latent Space.} Different colors represent latent space regions with different attribute scores. The boundary between two colored regions is an equipotential subspace. 
   %, where latent codes lying on this equipotential subspace all have the same semantic score. 
   Existing methods are represented by the trajectory \textcircled{1}, where latent code is shifted along a fixed direction throughout editing. Our method is represented by trajectory \textcircled{2}, where latent code is moved along location-specific directions.%, which are directly generate by Semantic Field.
   }
   \vspace{-0.3cm}
   \label{field}
\end{figure*}

Given an input image $\rmI \in \mathbb{R}^{3\times H \times W}$ and a pretrained GAN generator $G$, similar to previous latent space based manipulation methods \cite{interfacegan1, interfacegan2, enjoy_your_editing, pan20202d}, we need to firstly inverse the corresponding latent code $\vz \in \mathbb{R}^d$ such that $\rmI = G(\vz)$, and then find the certain vector $\vf_{z} \in \mathbb{R}^d$ which can change the attribute degree.
Note that adopting the same vector for all faces is vulnerable to identity change during editing, as different faces could have different $\vf_{z}$.
Thus, the vector should be \textit{location-specific}, \ie, the vector is not only unique to different identities but also varies during editing. 
% In other words, the vector is unique to each latent code.
Motivated by this, we propose to model the latent space as a continual ``semantic field'', \ie, a vector field that assigns a vector to each latent code. 

\noindent\textbf{Definition of Continual Semantic Field.}
For a latent code $\vz$ in the latent space, suppose its corresponding image $\rmI$ has a score $s$ for a certain attribute. By finding a proper vector $\vf_{z}$ and then adding the vector to $\vz$, the attribute score $s$ will be changed to $s^{\prime}$. Intuitively, the vector $\vf_{z}$ to increase the attribute score for the latent code $\vz$ is the gradient of $s$ with respect to $\vz$.

Mathematically, the attribute score is a scalar field, denoted as $S: \mathbb{R}^d \mapsto \mathbb{R}$. The gradient of attribute score field $S$ with respect to the latent code is a vector field, which we term as ``semantic field''. The semantic field $F: \mathbb{R}^d \mapsto \mathbb{R}^d$ can be defined as follows:
\begin{equation}
   F = \nabla S.
\end{equation}
For a specific latent code $\vz$, the direction of its semantic field vector $\vf_{z}$ is the direction in which the attribute score $s$ increases the fastest.

In the latent space, if we want to change the attribute score $s$ of a latent code $\vz$, all we need is to move $\vz$ along the latent direction in the semantic field. Due to the \textit{location-specific} property of the semantic field, the trajectory of changing the attribute score from $s_a$ to $s_b$ is curved. The formula for changing attribute score is expressed as:
\begin{equation}
   s_a + \int_{\vz_a}^{\vz_b} \vf_z \cdot \,d\vz = s_b,
   \label{equa_continual_field}
\end{equation}
where $\vz_a$ is the initial latent code and $\vz_b$ is the end point. 
As the semantic field is continuous and location-specific, continuous facial editing can be easily achieved by traversing the latent space along the semantic field line.

\noindent\textbf{Discretization of Semantic Field.} 
Though the attribute score field and semantic field in the real world are both continual, in practice, we need to discretize the continual field to approximate the real-world continual one. Thus, the discrete version of Eq.~(\ref{equa_continual_field}) can be expressed as:
\begin{equation}
s_a + \sum_{i = 1}^{N} \vf_{z_i} \cdot \Delta \vz_i  = s_b.
\end{equation}

The semantic field $F$ is implemented as a mapping network.
For a latent code $\vz$, we could obtain its corresponding semantic field vector via $\vf_{z} = F(\vz)$. 
Then one step of latent code shifting is achieved by:
\begin{align}
   \vz^{\prime} &= \vz + \alpha \vf_{z} \nonumber \\
              &= \vz + \alpha F(\vz), \label{eq:one_step}
\end{align}
where $\alpha$ is the step size, which is set to $\alpha=1$ in this work.
Since $\vf_{z}$ is supposed to change the attribute degree, the edited image $\rmI^{\prime} = G (\vz^{\prime})$ should have a different attribute score from the original image $\rmI = G(\vz)$.
During editing, we repeat Eq.~(\ref{eq:one_step}) until the desired attribute score is reached.

As illustrated in Fig.~\ref{field}, to train the mapping network so that it has the property of a semantic field, a pretrained fine-grained attribute predictor $P$ is employed to supervise the learning of semantic field. 
The predictor has two main functions: one is to push the output vector to change the attribute of interest in a correct direction, and the other is to keep the other irrelevant attributes unchanged.
Suppose we have $k$ attributes in total. The fine-grained attributes of the original image can be denoted as $(a_{1}, a_{2}, ..., a_{i}, ..., a_{k})$, where $a_{i} \in \{0,1,...,C\}$ are the discrete class labels indicating the attribute degree. 
When we train the semantic field for the $i$-th attribute, the target attributes labels $y$ of the edited image $\rmI^{\prime}$  
should be $(a_{1}, a_{2}, ..., a_{i} + 1, ..., a_{k})$. With the target attribute labels, we can optimize the desired semantic field using the cross-entropy loss, %of the pretrained predictor. The predictor loss $L_{pred}$ 
then the predictor loss $L_{pred}$ is expressed as follows:
\vspace{-0.4cm}
\begin{equation}
\vspace{-0.1cm}
   L_{pred} = - \sum_{i=1}^{k} \sum_{c=0}^{C}y_{i,c}log(p_{i, c}),
\end{equation}
where $C$ denotes the number of fine-grained classes, $y_{i, c}$ is the binary indicator with respect to the target class, and $p_{i, c}$ is the softmax output of predictor $P$, \ie, $p = P(\rmI^\prime)$.

As the \textit{location-specific} property of the semantic field allows different identities to have different vectors, we further introduce an identity keeping loss \cite{wang2021gfpgan, taigman2016unsupervised} to better preserve the face identity when shifting the latent codes along the semantic field.
Specifically, we employ an off-the-shelf face recognition model to extract discriminative features, and the extracted features during editing should be as close as possible.
The identity keeping loss $L_{id}$ is defined as follows:
\begin{equation}
   L_{id} = \left \| Face(\rmI^{\prime}) - Face(\rmI) \right \|_{1},
\end{equation}
where $Face(\cdot)$ is the pretrained face recognition model \cite{deng2019arcface}.

Moreover, to avoid unrealistic artifacts in edited images, 
we could further leverage the pretrained discriminator $D$ coupled with the face generator as follows:
\begin{equation}
   L_{disc} = - D(\rmI^\prime).
\end{equation}

To summarize, we use the following loss functions to supervise the learning of semantic field:
\begin{equation}
   L_{total} = \lambda_{pred} L_{pred} + \lambda_{id} L_{id} + \lambda_{disc} L_{disc},
\end{equation}
where $\lambda_{pred}$, $\lambda_{id}$ and $\lambda_{disc}$ are weights for predictor loss, identity keeping loss and discriminator loss respectively.

\begin{figure*}
   \begin{center}
      \includegraphics[width=\linewidth]{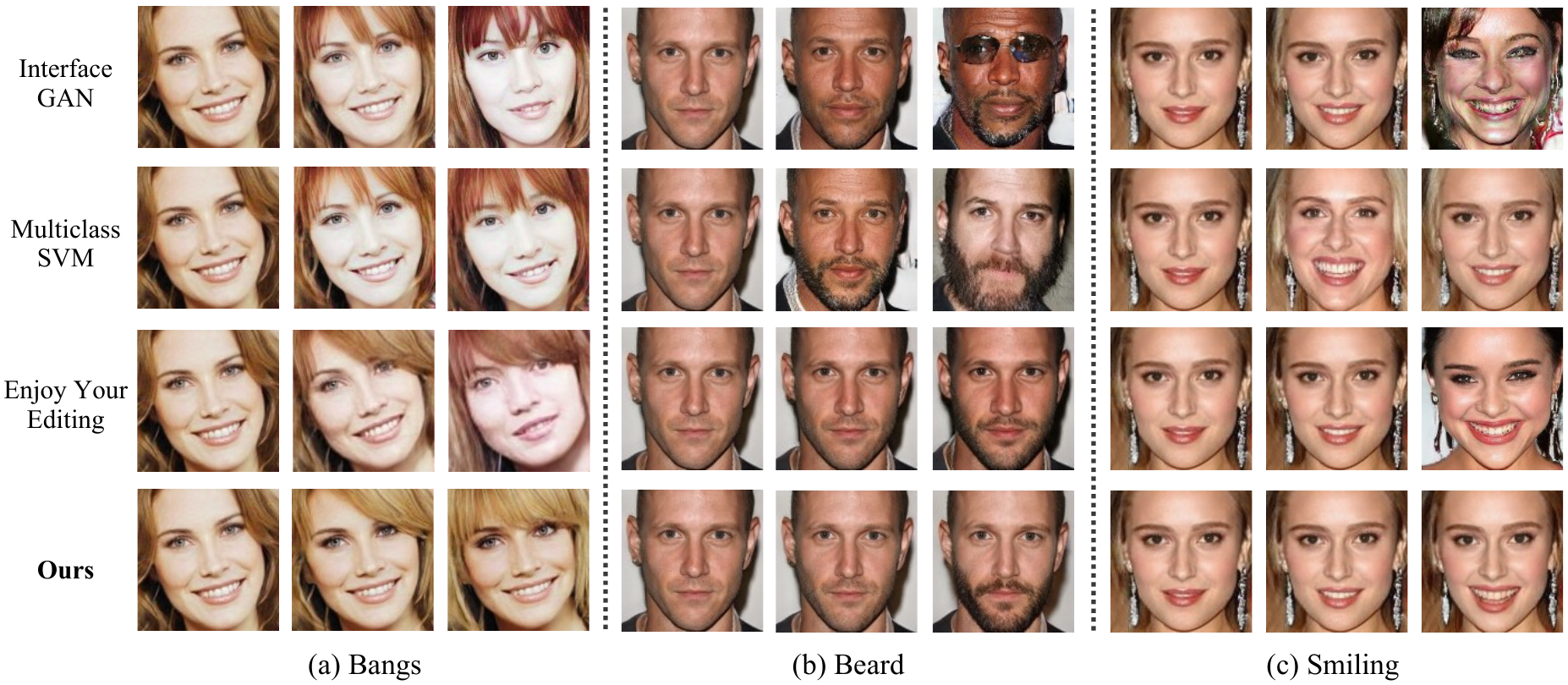}
   \end{center}
   \vspace{-0.7cm}
   \caption{\textbf{Qualitative Comparison.} We compare our approach with InterfaceGAN, Multiclass SVM and Enjoy Your Editing. Our editing results are more realistic. Besides, our method is less likely to change the identity and other attributes.}
   \vspace{-0.3cm}
   \label{qualitative}
\end{figure*}

\subsection{System Feedback}

The system \textit{Talk} module provides natural language feedback as follows:
\begin{equation}
    feedback_{t} = Talk(feedback_{t-1}, \vr, \vs, \ve_{r}, \vh),
\end{equation}
where $\vr$ is the user request, $\vs$ is the current system state, $\ve_{r}$ is the editing encoding, and $\vh$ is the editing history.

The feedback provided by the system comes from one of the three categories: 
\textbf{1)} checking if the attribute degree of the edited image meets users' expectations,
\textbf{2)} providing alternative editing suggestions or options, and
\textbf{3)} asking for further user instructions.

%% file: section/experiment.tex
\begin{table*}[]
\begin{center}
\caption{\textbf{Quantitative Comparisons.} We report Identity / Attribute preservation metrics. A lower identity score (smaller feature distance) means the identity is better preserved, and a lower attribute score (smaller cross-entropy) means the irrelevant attributes are less changed. Our method has a superior performance in terms of identity and attribute preservation.}
\vspace{-6pt}
\begin{tabular}{l|ccccc}
\Xhline{1pt}
\textbf{Methods}    & \textbf{Bangs} & \textbf{Eyeglasses} & \textbf{Beard} & \textbf{Smiling} & \textbf{Young} \\ \Xhline{1pt}
InterfaceGAN        & 0.7621 / 0.7491 & 0.7831 / 1.1904 & 1.0213 / 1.6458 & 0.9158 / 0.9030 & 0.7850 / 1.4169 \\ \hline
Multiclass SVM      & 0.7262 / 0.5387 & 0.6967 / 0.9046 & 1.1098 / 1.7361 & 0.7959 / 0.8676 & 0.7610 / 1.3866 \\ \hline
Enjoy Your Editing  & 0.6693 / 0.4967 & 0.7341 / 0.9813 & 0.8696 / 0.7906 &  0.6639 / 0.5092 & 0.7089 / 0.5734 \\ \hline \hline
Talk-to-Edit (Ours) & 0.6047 / 0.3660 & \textbf{0.6229} / 0.7720 & 0.8324 / 0.6891 & 0.6434 / 0.5028 & 0.6309 / 0.4814 \\ \hline
Talk-to-Edit (Ours) *  & \textbf{0.5276} / \textbf{0.2902} & 0.6670 / \textbf{0.6345} & \textbf{0.7634} / \textbf{0.5425} & \textbf{0.4580} / \textbf{0.3573} & \textbf{0.6234} / \textbf{0.2731} \\\Xhline{1pt}
\end{tabular}
\vspace{-0.3cm}
\label{quantitative}
\end{center}
\end{table*}

\begin{figure*}
   \begin{center}
      \includegraphics[width=0.9\linewidth]{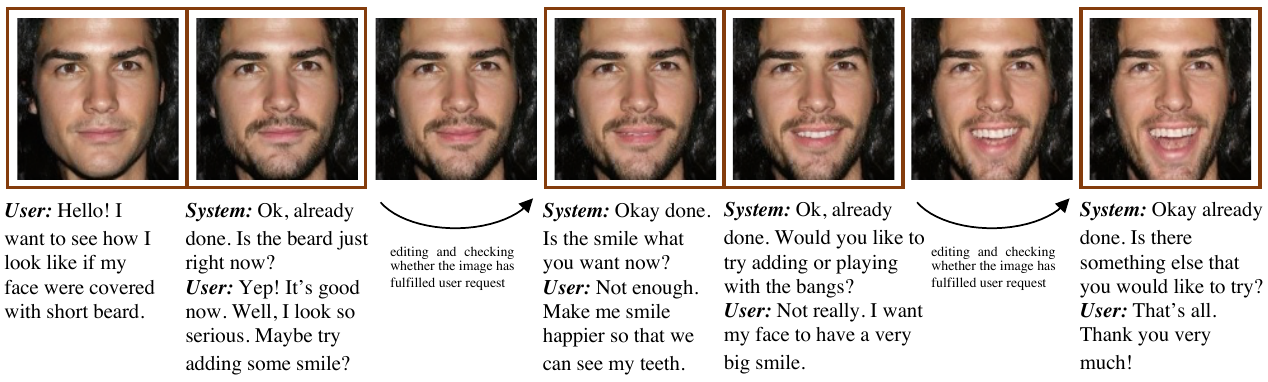}
   \end{center}
   \vspace{-0.6cm}
   \caption{\textbf{Results of dialog-based facial editing.} The whole process is driven by the dialog between the user and the system.}
   \vspace{-0.3cm}
   \label{dialog_demo}
\end{figure*}

\section{Experiments}

\noindent\textbf{Evaluation Datasets.}
We synthesize the evaluation dataset by sampling latent codes from the StyleGAN pretrained on CelebA dataset \cite{celeba}. 
Using latent codes, we then generate corresponding images.
When comparing with other latent space based manipulation methods, we use the latent code for editing directly to avoid the error introduced by GAN-inversion methods. 
Considering computation resources, we compare our method with baselines on $128 \times 128$ images.

\noindent\textbf{Evaluation Metrics.}
We evaluate the performance of facial editing methods in terms of 
identity and attribute preservation as well as the photorealism of edited images. 
To evaluate the identity preservation, we extract the features of the images before and after editing with FaceNet \cite{schroff2015facenet}, and compute their euclidean distance.
As for the irrelevant attribute preservation, we use a retrained attribute predictor to output a cross-entropy score indicating whether the predicted attribute is consistent with its ground-truth label. 

Apart from the aforementioned metrics, we also conduct a user study.
Two groups of editing results (one is our result, the other is another method) are provided to participants. The participants are supposed to compare two groups of editing images and then choose the more suitable group for each of the following questions: 1) \textit{Which group of images is more visually realistic?} 2) \textit{Which group of images has more continuous changes?} 3) \textit{After editing, which group of images better preserves the identity?}

\begin{figure*}
  \begin{center}
      \includegraphics[width=0.95\linewidth]{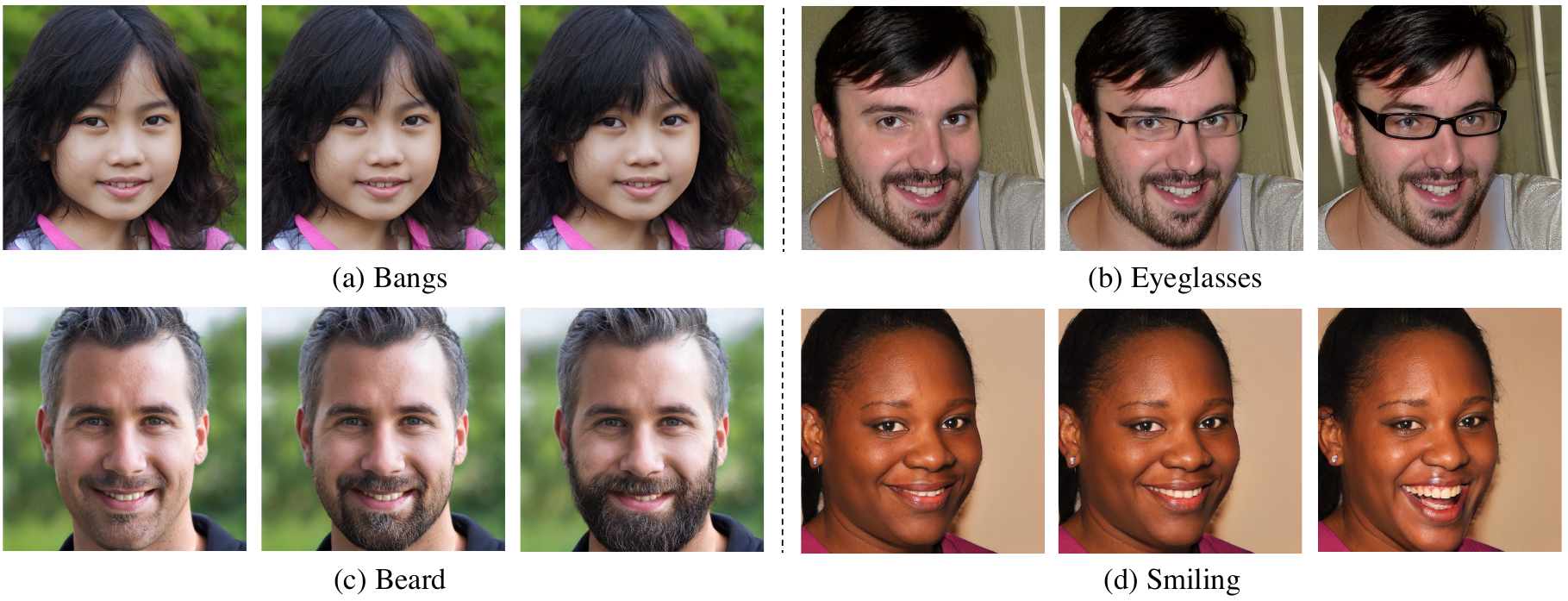}
  \end{center}
  \vspace{-0.5cm}
  \caption{\textbf{High-Resolution Image Editing.} Our method can be generalized to $1024 \times 1024$ images.} 
  \vspace{-0.2cm}
  \label{HQ_results}
\end{figure*}

\begin{figure*}
   \begin{center}
      \includegraphics[width=0.95\linewidth]{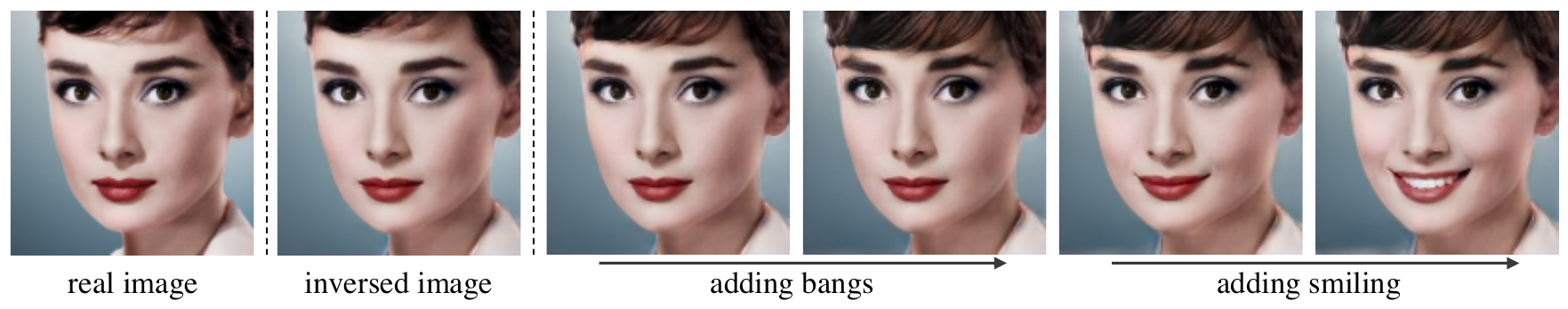}
   \end{center}
   \vspace{-0.5cm}
   \caption{\textbf{Real Image Editing.} Given a real image, we first inverse the image and find its corresponding latent code in latent space. We firstly add bangs and then add smiling.}
   \vspace{-0.4cm}
   \label{inversion}
\end{figure*}

\begin{figure}
   \begin{center}
      \includegraphics[width=\linewidth]{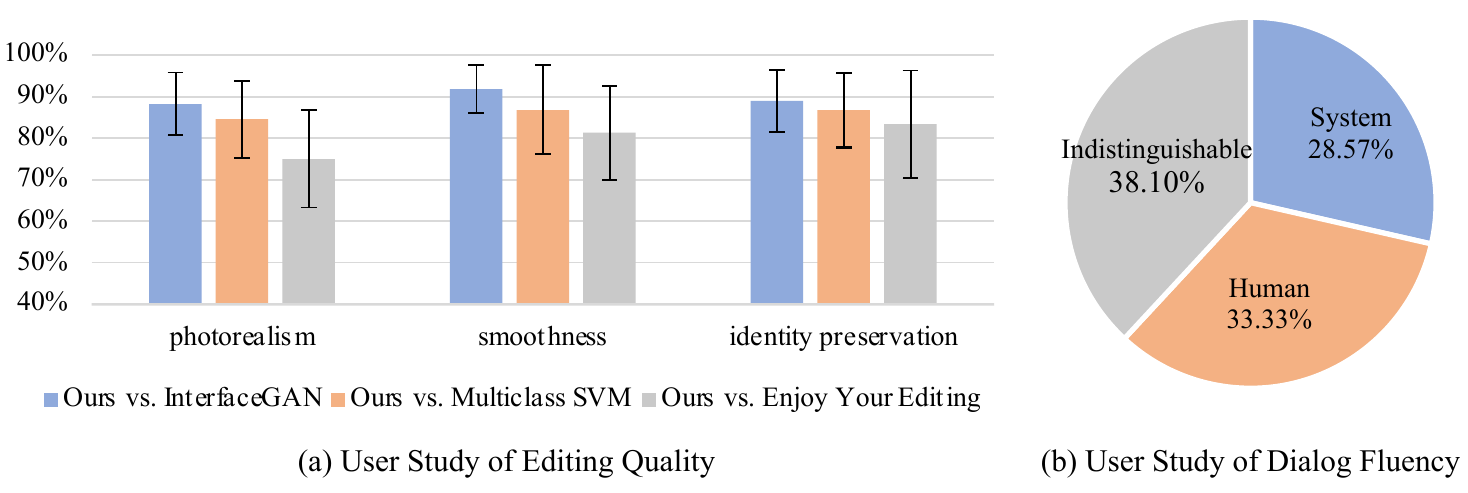}
   \end{center}
  \vspace{-0.6cm}
   \caption{\textbf{User Study.} (a) The percentage of participants favoring our results against existing methods. Our results are preferred by the majority of participants. (b) Over half of the participants think the system feedback is natural.}
   \vspace{-0.4cm}
   \label{user_study}
\end{figure}

\begin{figure}
   \vspace{0.1cm}
   \begin{center}
      \includegraphics[width=\linewidth]{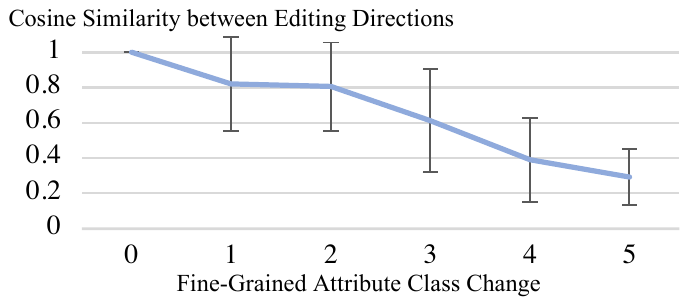}
   \end{center}
   \vspace{-0.5cm}
  \caption{\textbf{Cosine Similarity.} We compute the average cosine similarity between the initial direction and directions of later steps. As the attribute class changes, the cosine similarity decreases, indicating that the editing trajectories for most facial images are curved.}
   \vspace{-0.5cm}
   \label{cosine_similartiy}
\end{figure}

\subsection{Comparison Methods}

\noindent\textbf{InterfaceGAN.}
InterfaceGAN \cite{interfacegan2} uses a single direction to perform continuous editing. The direction is obtained by computing the normal vector of the binary SVM boundary.

\noindent\textbf{Multiclass SVM.} 
We further propose an extended version of InterfaceGAN, named Multiclass SVM, where fine-grained labels are used to get multiple SVM boundaries. During the editing, directions will be constantly switched.

\noindent\textbf{Enjoy Your Editing.}
Enjoy your editing \cite{enjoy_your_editing} learns a mapping network to generate an identity-specific direction, and it keeps fixed during editing for one identity.

\subsection{Quantitative Evaluation}

\noindent\textbf{Identity/Attribute Preservation.}
To fairly compare the continuous editing results with existing methods, we produce our results purely based on semantic field manipulation and language is not involved. We compute the identity preservation and attribute preservation scores for the editing results of baseline methods. Table \ref{quantitative} shows the quantitative comparison results. Our method achieves the best identity and attribute preservation scores.

\noindent\textbf{Ablation Study.}
The \textit{location-specific} property of semantic field has the following two indications: 1) the trajectory to edit one identity might be a curve instead of a straight line; 2) the editing trajectories are unique to individual identities. The superiority over InterfaceGAN and Enjoy Your Editing validates that the curved trajectory is vital for continuous editing and we will provide further analysis in Section~\ref{sec:analysis}. Compared to Multiclass SVM, our results confirm the necessity of different directions for different identities.

\subsection{Qualitative Evaluation}

\noindent\textbf{Visual Photorealism.}
Qualitative comparisons are shown in Fig. \ref{qualitative}. The results of our method displayed are edited on W+ space. Our proposed method is less likely to generate artifacts compared to previous methods. Besides, when the edited attribute comes to higher degrees, our method can still generate plausible editing results while keeping the identity unchanged.

\noindent\textbf{User Study.}
We conduct a user study, where users are asked the aforementioned questions and they need to choose the better images. A total number of 27 participants are involved and they are required to compare 25 groups of images. We mix the editing results of different attributes together in the user study. The results of user study are shown in Fig. \ref{user_study} (a). The results indicate that the majority of users prefer our proposed method in terms of image photorealism, editing smoothness, and identity preservation.

\makeatletter{\renewcommand*{\@makefnmark}{}
  \footnotetext{$^*$ edits on W+ space. Others edit on Z space.}\makeatother}

\noindent\textbf{Dialog Fluency.}
In Fig. \ref{dialog_demo}, we show a dialog example, where the system is asked to add beard for the young guy in the picture. 
After adding the beard into a desired one, the system then continues to edit the smile as required by the user. The system could talk to the user smoothly in the whole dialog.
To further evaluate the fluency of dialog, we invite seven participants to compare six pairs of dialog. In each pair of dialog, one is generated by the system, and the other is revised by a human. Participants need to decide which one is more natural or if they are indistinguishable. The results are shown in Fig. \ref{user_study} (b). Over half of the participants think the system feedback is natural and fluent.

\subsection{Further Analysis}
\label{sec:analysis}

\noindent\textbf{High-Resolution Facial Editing.} 
Since our editing method is a latent space manipulation based method, it can be extended to images with any resolutions as long as the pretrained GAN is available. Apart from editing results on $128 \times 128$ images shown in previous parts, we also provide some $1024 \times 1024$ resolution editing results in Fig. \ref{HQ_results}.

\noindent\textbf{Location-specific Property of Semantic Field.}
When traversing the semantic field, the trajectory to change the attribute degree is determined by the curvature at each step, and thus it is curved. 
To further verify this hypothesis, we randomly sample 100 latent codes and then continuously add eyeglasses for the corresponding $1024 \times 1024$ images.
For every editing direction, we compute its cosine similarity with the initial direction. 
The average cosine similarity against the attribute class change is plotted in Fig. \ref{cosine_similartiy}.
We observe that the cosine similarity tends to decrease as the attribute class change increases. 
It confirms that the editing direction could constantly change according to its current location, and thus the location-specific property is vital for continuous editing and identity preservation.

\noindent\textbf{Real Image Editing.} 
In Fig. \ref{inversion}, we show an example of real image editing results. The image is firstly inversed by the inversion method proposed by Pan \etal \cite{pan2020exploiting}. The inversion process would finetune the weight of StyleGAN, and we observe that the trained semantic field still works.

%% file: section/conclusion.tex
\section{Conclusion}

In this paper, we present a dialog-based fine-grained facial editing system named \textbf{Talk-to-Edit}. The desired facial editing is driven by users' language requests and the system is able to provide feedback to users to make the facial editing more feasible. 
By modeling the non-linearity property of the GAN latent space using semantic field, our proposed method is able to deliver more continuous and fine-grained editing results.
We also contribute a large-scale visual-language facial attribute dataset named \textbf{CelebA-Dialog}, which we believe would be beneficial to fine-grained and language driven facial editing tasks.
In future work, the performance of real facial image editing can be further improved by incorporating more robust GAN-inversion methods and adding stronger identity keeping regularization. 
We also hope to deal with more complex text requests by leveraging advanced pretrained language models.

\noindent\textbf{Acknowledgement}.
This study is supported by NTU NAP, MOE AcRF Tier 1 (2021-T1-001-088), and under the RIE2020 Industry Alignment Fund – Industry Collaboration Projects (IAF-ICP) Funding Initiative, as well as cash and in-kind contribution from the industry partner(s).

%% file: section/iccv_supp.tex
\onecolumn
\appendix
\section*{Supplementary}
\renewcommand\thesection{\Alph{section}}
\renewcommand\thefigure{A\arabic{figure}}
\renewcommand\thetable{A\arabic{table}}

In this supplementary file, we will explain the detailed annotation definition of CelebA-Dialog Dataset in Section \ref{dataset_definition}. Then we will introduce implementation details in Section \ref{implementation_details}. 
In Section \ref{further_experiment}, we will give more detailed explanations on experiments, including evaluation dataset, evaluation metrics and implementation details on comparison methods. Then we provide more visual results in Section \ref{qualitative}. 
Finally, we will discuss failure cases in Section \ref{failur_discuss}.

\section{CelebA-Dialog Dataset Annotations}
\label{dataset_definition}

For each image, fine-grained attribute annotations and textual descriptions are provided. In Table \ref{bangs_annotation} - \ref{age_annotation}, we give detailed definitions of fine-grained attribute annotations. With each fine-grained attribute label, we also provide an example image and its corresponding textual description. 

\begin{center}
    \makeatletter\def\@captype{table}\makeatother
    \caption{\textbf{Annotation Definition and Examples of \textit{Bangs} Attribute.}}
	\begin{tabular}{c|l|l}
		\Xhline{1pt}
		\textbf{Attribute Degree}   & \multicolumn{1}{c|}{\textbf{Fine-Grained Definition}} & \textbf{Examples}  \\
		\Xhline{1pt}
		0 & without bangs, full forehead exposed & \begin{minipage}{0.1\textwidth}
		    \vspace{3pt}
            \includegraphics[width=0.8\textwidth]{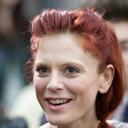}
        \end{minipage} \textit{The lady has no bangs.} \\
		1 & very short bangs, 80\% forehead exposed & \begin{minipage}{0.1\textwidth}
		    \vspace{3pt}
            \includegraphics[width=0.8\textwidth]{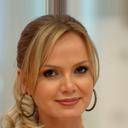} 
        \end{minipage} \textit{She has very short bangs covering her forehead.}\\
		2 & short bangs, 60\% forehead exposed &
		\begin{minipage}{0.1\textwidth}
		    \vspace{3pt}
            \includegraphics[width=0.8\textwidth]{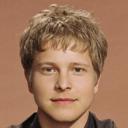} 
        \end{minipage} \makecell[l]{\textit{The man has short bangs that cover a small} \\ \textit{portion of the forehead.} }
		 \\
		3 & medium bangs, 40\% forehead exposed & \begin{minipage}{0.1\textwidth}
		    \vspace{3pt}
            \includegraphics[width=0.8\textwidth]{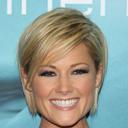} 
        \end{minipage} \textit{The woman has bangs of medium length.} \\
		4 & long bangs, 20\% forehead exposed & \begin{minipage}{0.1\textwidth}
		    \vspace{3pt}
            \includegraphics[width=0.8\textwidth]{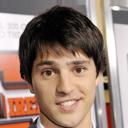} 
        \end{minipage} \textit{The guy has long bangs.} \\
		5 & extremely long bangs, all forehead covered & \begin{minipage}{0.1\textwidth}
		    \vspace{3pt}
            \includegraphics[width=0.8\textwidth]{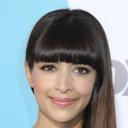}
            \vspace{3pt}
        \end{minipage} \textit{The woman has bangs that cover the eyebrows.} \\
		\Xhline{1pt}
	\end{tabular}
	\label{bangs_annotation}
\end{center}

\newpage

\begin{center}
    \makeatletter\def\@captype{table}\makeatother
    \caption{\textbf{Annotation Definition and Examples of \textit{Eyeglasses} Attribute.}}
	\begin{tabular}{c|l|l}
		\Xhline{1pt}
		\textbf{Attribute Degree}   & \multicolumn{1}{c|}{\textbf{Fine-Grained Definition}} & \textbf{Examples}  \\
		\Xhline{1pt}
		0 & no eyeglasses & \begin{minipage}{0.1\textwidth}
		    \vspace{3pt}
            \includegraphics[width=0.8\textwidth]{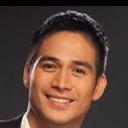} 
        \end{minipage} \textit{The man doesn't wear eyeglasses.} \\
		1 & \makecell[l]{eyeglasses with a very thin metal frame or \\ no frame.}   & \begin{minipage}{0.1\textwidth}
		    \vspace{3pt}
            \includegraphics[width=0.8\textwidth]{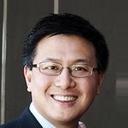} 
        \end{minipage} \textit{He wears a pair of rimless eyeglasses.} \\
		2 & \makecell[l]{eyeglasses with a thicker metal frame \\ or thinner plastic frame.}   & \begin{minipage}{0.1\textwidth}
		    \vspace{3pt}
            \includegraphics[width=0.8\textwidth]{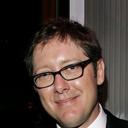} 
        \end{minipage} \textit{His eyeglasses have a thin frame.} \\
		3 & eyeglasses with a thick frame or plastic frame & \begin{minipage}{0.1\textwidth}
		    \vspace{3pt}
            \includegraphics[width=0.8\textwidth]{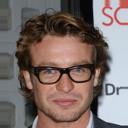} 
        \end{minipage} \textit{The man wears eyeglasses of a thick frame.} \\
		4 & sunglasses with a thin frame & \begin{minipage}{0.1\textwidth}
		    \vspace{3pt}
            \includegraphics[width=0.8\textwidth]{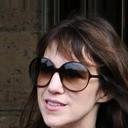} 
        \end{minipage} \makecell[l]{\textit{The lady wears a pair of sunglasses with} \\ \textit{a thin frame.}} \\
		5 & sunglasses with a thick frame & \begin{minipage}{0.1\textwidth}
		    \vspace{3pt}
            \includegraphics[width=0.8\textwidth]{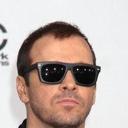}
            \vspace{3pt}
        \end{minipage} \makecell[l]{\textit{He wears sunglasses that have a thick} \\ \textit{frame.}} \\
		\Xhline{1pt}
	\end{tabular}
	\label{eyeglasses_annotation}
\end{center}

\begin{center}
    \makeatletter\def\@captype{table}\makeatother
    \caption{\textbf{Annotation Definition and Examples of \textit{Beard} Attribute.}}
	\begin{tabular}{c|l|l}
		\Xhline{1pt}
		\textbf{Attribute Degree}   & \multicolumn{1}{c|}{\textbf{Fine-Grained Definition}} & \textbf{Examples}  \\
		\Xhline{1pt}
		0 & no beard & \begin{minipage}{0.1\textwidth}
		    \vspace{3pt}
            \includegraphics[width=0.8\textwidth]{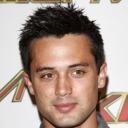} 
        \end{minipage} \textit{There is no beard on his face.} \\
		1 & with a shaved beard, very short in length & \begin{minipage}{0.1\textwidth}
		    \vspace{3pt}
            \includegraphics[width=0.8\textwidth]{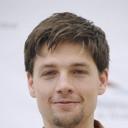} 
        \end{minipage} \makecell[l]{\textit{The man's face is covered with the short} \\ \textit{pointed beard.}} \\
		2 & with a beard that hasn't been shaved for a while & \begin{minipage}{0.1\textwidth}
		    \vspace{3pt}
            \includegraphics[width=0.8\textwidth]{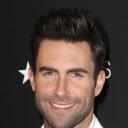} 
        \end{minipage} \textit{He has a short beard.} \\
		3 & with a deliberate beard of medium length & \begin{minipage}{0.1\textwidth}
		    \vspace{3pt}
            \includegraphics[width=0.8\textwidth]{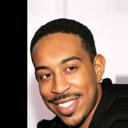} 
        \end{minipage} \makecell[l]{\textit{His face is covered with beard of medium} \\ \textit{length.}} \\
		4 & with a long but tiny beard & \begin{minipage}{0.1\textwidth}
		    \vspace{3pt}
            \includegraphics[width=0.8\textwidth]{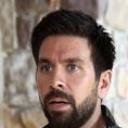} 
        \end{minipage} \textit{He has a long but tiny beard.} \\
		5 & with a very bushy, long and untidy beard & \begin{minipage}{0.1\textwidth}
		    \vspace{3pt}
            \includegraphics[width=0.8\textwidth]{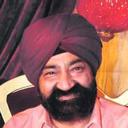}
            \vspace{3pt}
        \end{minipage} \textit{He has a very long beard.} \\
		\Xhline{1pt}
	\end{tabular}
	\label{beard_annotation}
\end{center}

\vspace{-5cm}

\begin{center}
    \makeatletter\def\@captype{table}\makeatother
    \caption{\textbf{Annotation Definition and Examples of \textit{Smiling} Attribute.}}
	\begin{tabular}{c|l|l}
		\Xhline{1pt}
		\textbf{Attribute Degree}   & \multicolumn{1}{c|}{\textbf{Fine-Grained Definition}} & \textbf{Examples}  \\
		\Xhline{1pt}
		0 & no smile on face & \begin{minipage}{0.1\textwidth}
		    \vspace{3pt}
            \includegraphics[width=0.8\textwidth]{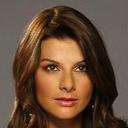} 
        \end{minipage} \textit{The woman looks serious.} \\
		1 & smile without teeth exposed & \begin{minipage}{0.1\textwidth}
		    \vspace{3pt}
            \includegraphics[width=0.8\textwidth]{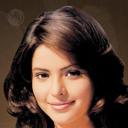} 
        \end{minipage} \textit{She has a tight-lipped smile on her face.} \\
		2 & smile with some teeth exposed & \begin{minipage}{0.1\textwidth}
		    \vspace{3pt}
            \includegraphics[width=0.8\textwidth]{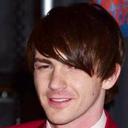} 
        \end{minipage} \makecell[l]{\textit{He smiles with the corners of the mouth} \\ \textit{curved up and some teeth exposed.}} \\
		3 & laughing with the whole row of teeth exposed & \begin{minipage}{0.1\textwidth}
		    \vspace{3pt}
            \includegraphics[width=0.8\textwidth]{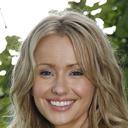} 
        \end{minipage} \textit{She has a beaming face.} \\
		4 & laughing with mouth moderately open & \begin{minipage}{0.1\textwidth}
		    \vspace{3pt}
            \includegraphics[width=0.8\textwidth]{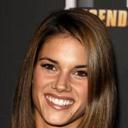} 
        \end{minipage} \textit{There is a big smile on her face.} \\
		5 & exaggerated laughing with mouth widely open & \begin{minipage}{0.1\textwidth}
		    \vspace{3pt}
            \includegraphics[width=0.8\textwidth]{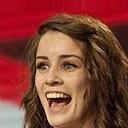}
            \vspace{3pt}
        \end{minipage} \makecell[l]{\textit{The woman smiles with the mouth widely} \\ \textit{open.}}\\
		\Xhline{1pt}
	\end{tabular}
	\label{smiling_annotation}
\end{center}

\begin{center}
    \makeatletter\def\@captype{table}\makeatother
    \caption{\textbf{Annotation Definition and Examples of \textit{Young} Attribute.}}
	\begin{tabular}{c|l|l}
		\Xhline{1pt}
		\textbf{Attribute Degree}   & \multicolumn{1}{c|}{\textbf{Fine-Grained Definition}} & \textbf{Examples}  \\
		\Xhline{1pt}
		0 & under 15 years old with childish face & \begin{minipage}{0.1\textwidth}
		    \vspace{3pt}
            \includegraphics[width=0.8\textwidth]{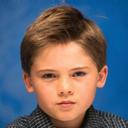} 
        \end{minipage} \textit{The person in the picture is under 15 years old.} \\
		1 & 15-30 years old, adolescent & \begin{minipage}{0.1\textwidth}
		    \vspace{3pt}
            \includegraphics[width=0.8\textwidth]{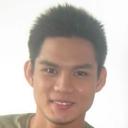} 
        \end{minipage} \textit{He is at the age of adolescent.} \\
		2 & 30-40 years old, mature youth & \begin{minipage}{0.1\textwidth}
		    \vspace{3pt}
            \includegraphics[width=0.8\textwidth]{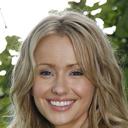} 
        \end{minipage} \textit{The woman is in the thirties.} \\
		3 & 40-50 years old, middle-aged & \begin{minipage}{0.1\textwidth}
		    \vspace{3pt}
            \includegraphics[width=0.8\textwidth]{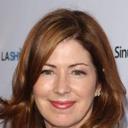} 
        \end{minipage} \textit{She looks like a middle age one.} \\
		4 & 50-60 years old & \begin{minipage}{0.1\textwidth}
		    \vspace{3pt}
            \includegraphics[width=0.8\textwidth]{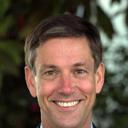} 
        \end{minipage} \textit{He is at the age of his fifties.} \\
		5 & over 60 years old & \begin{minipage}{0.1\textwidth}
		    \vspace{3pt}
            \includegraphics[width=0.8\textwidth]{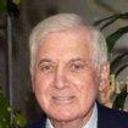} 
            \vspace{3pt}
        \end{minipage} \textit{The man is very old.} \\
		\Xhline{1pt}
	\end{tabular}
	\label{age_annotation}
\end{center}

% \vspace{-1.2cm}

\section{Implementation Details}
\label{implementation_details}

\subsection{User Request Understanding} 

The language encoder $E$ has three components: 1) a learnable 300-D word embedding; 2) a two-layer LSTM with cell size of 1024; 3) fully-connected layers following the LSTM to generate the editing encoding $\ve_{r}$. 
The learning rate is set as $10^{-3}$, the batch size is 2048, and the Adam optimizer \cite{kingma2015adam} is adopted.

Commonly, users' editing requests could be roughly classified into three major types: 
\textbf{1)} Describe the attribute and specify the target degree, \eg, \textit{Let's try extremely long bangs that cover the entire forehead.} 
\textbf{2)} Describe the attribute of interest and indicate the relative degree of change, \eg, \textit{The bangs can be slightly longer.} 
\textbf{3)} Describe the attribute and only the editing direction without specifying the degree of change, \eg, \textit{Let's make the bangs longer.}
Since the types of facial editing requests are relatively fixed, we use template-based text generation methods to form a pool of editing requests. The request pool is used to train the language encoder. 
We prepare more than 300 request templates with diverse sentence patterns. A pool of synonymous words is used to enrich the user request templates. We generate 10,000 user requests in total. For each generated request, we provide their corresponding hard labels to train the language encoder $E$. 

The editing encoding $\ve_{r}$ generated by the language encoder $E$ is implemented as hard labels containing the following information: (1) request type, (2) the attribute of interest, (3) the editing direction, and (4) the change of degree. In practice, the same user request could be interpreted differently depending on the dialog context. For example, simply saying \textit{``Yes"} has different meanings under different scenarios. If the system makes a suggestion \textit{``Do you want to make the bangs longer?"}, by replying \textit{``Yes"}, the user means to make the bangs longer. However, if the system asks if the desired effect is achieved in the previous round, \textit{``Yes"} means the editing is satisfactory in this context. Therefore, multiple language encoders are needed to parse the user request under different dialog context. During training, the weights of word embedding and LSTM are shared across different language encoders. The current system feedback decides which language encoder would be used.

We track the dialog-based editing system using a finite-state machine. The editing system is in one of the four states at any moment: \textbf{1)} \textit{start}, that is, the first round of dialog,
\textbf{2)} \textit{edit}, where the system performs editing in the current round of dialog,
\textbf{3)} \textit{no edit}, where the system does not edit the image and wait for further instructions from the user.
and \textbf{3)} \textit{end}, where the system ends the conversation upon the user's request.

\subsection{Semantic Field} 

The training of semantic field requires the following pretrained models: fine-grained attribute predictor $P$, face recognition model $Face$, StyleGAN generator $G$ and discriminator $D$.
The fine-grained attribute predictor $P$ is pretrained on CelebA-Dialog dataset using our fine-grained attribute labels with a multi-class cross-entropy loss. 
StyleGAN $G$ and its corresponding discriminator $D$ are trained on CelebA dataset \cite{celeba} and FFHQ dataset \cite{karras2019style} for $128 \times 128$ and $1024 \times 1024$ facial images respectively. As for the $Face$ Model, we use the off-the-shelf ArcFace model \cite{deng2019arcface} trained on LFW dataset \cite{LFWTech, LFWTechUpdate}. 

Since the pretrained StyleGAN has the mode collapse problem, during the training of semantic field, we need to sample the training latent codes such that all fine-grained attribute classes are more balancedly distributed. 
The mapping network of semantic field $F$ is composed of 8 fully-connected (FC) layers with dimension 512. Except for the last FC layer, each FC layer is followed by a leaky ReLU with slope 0.2.
The learning rate for training the semantic field is $10^{-4}$, batch size is set as 32, and Adam optimizer \cite{kingma2015adam} is adopted. 

We also provide editing results on W+ sapce. When editing on W+ space, to enforce the field vector to be a valid vector that would not make the edited latent code fall into the outlier region of pretrained StyleGAN latent space, we adopt a regularization method proposed by Pan \etal \cite{pan20202d}. The latent code is updated as follows:
\begin{align}
    \vz^{\prime} &= \vz + \alpha (M(\vf_{z}) - M(\vzero)) \nonumber \\
                &= \vz + \alpha (M(F(\vz)) - M(\vzero)),
\label{latent_update}
\end{align}
where $M(\cdot)$ denotes the mapping network of StyleGAN and $F(\cdot)$ denotes the mapping network of the semantic field.

Besides, we found that the last few layers of latent codes of W+ space control the low-level features of a facial images, such as color, brightness, illuminations and etc. During facial editing, we need to keep these factors fixed. Therefore, when updating latent codes using Eq.~(\ref{latent_update}), we only update first $k$ layers of latent codes. We empirically set $k$ as $8$ for $128 \times 128$ images and $10$ for $1024 \times 1024$ images.

\subsection{System Feedback}
After editing an image, the \textit{Talk} module will provide a feedback, which belongs to one of the following categories: 1) checking whether the attribute degree is satisfying, in order to achieve fine-grained editing desired by the user. For example, after the user requests to make the bangs longer, the system could give the following feedback, \eg, \textit{“Are the bangs now of the length you like?"}. If in the previous round the user agrees on to edit an attribute suggested by system but does not specify the editing direction, then the system feedback will always be checking with the user about the attribute degree.
2) providing further editing suggestions, \eg, \textit{“Do you want to try manipulating the age?"} In order to let the user fully explore possible manipulation options, the system tends not to suggest editing an attribute that has been edited before. If there exist a larger number of attributes not edited by user yet, then there is a higher probability for the system to make a suggestion,
and 
3) asking for user instructions , \eg, \textit{“Ok, what's next?"}.

We sample a sentence from a pool of templates of the chosen feedback category, and randomly replace phrases using a predefined pool of synonyms to extend the language richness.
We observe that this simple design can provide meaningful feedback to some extent.

\section{Further Explanations on Experimental Details}
\label{further_experiment}

\subsection{Evaluation Dataset}

The latent code used for evaluation is formed by sampling latent codes from StyleGAN pretrained on CelebA datasets \cite{celeba}. Though the StyleGAN has demonstrated its powerful generative ability in facial image generation, some synthesized images are still of low quality. Thus, we need to manually filter the bad images with artifacts out. For the evaluation dataset of the eyeglasses attribute, latent codes whose corresponding images with degree above 0 are selected, as we observe that for all methods (including baselines) editing images with degree 0 would often make the latent code fall into out-of-distribution regions (corresponding images become artifacts). To avoid the error introduced by the aforementioned issue, we only use latent codes with attribute degree above 0. When constructing the evaluation dataset of the beard attribute, we adopt the same strategy so that images with females are excluded (No females would have beard attribute degree larger than 0).

\subsection{Evaluation Metrics}
We employ Identity Preservation Metrics and Attribute Preservation Metrics to evaluate the identity and attribute preservation respectively. Here we explain these two metrics in detail.

\noindent\textbf{Identity Preservation Metrics.} We use the off-the-shelf face model \textit{FaceNet} \cite{schroff2015facenet} to extract features for images before and after editing. Then we compute the euclidean distance between features of the edited facial images and the feature of the original facial image. The identity preservation metrics is expressed as follows:
\begin{equation}
    \mathrm{Identity Preservation} = \frac{1}{N} \sum_{i=1}^{N}\left \| FaceNet(I_i) - FaceNet(I_0) \right \|_2,
\end{equation}
where $I_0$ is the original image, $I_i$ are edited images, and $N$ is the total number of edited images.

\noindent\textbf{Attribute Preservation Metrics.} We retrain a attribute predictor $P^{\prime}$ (different from the one we use for training), and use the retrained predictor to output cross entropy score. The attribute preservation metrics is defined as follows:
\begin{equation}
    \mathrm{Attribute Preservation} = - \frac{1}{N} \sum_{i=1}^{N} \sum_{j=1, j \neq m}^{k} \sum_{c=0}^{C}y_{j,c}log(p_{j, c}^{\prime}),
\end{equation}
where $N$ is the total number of edited images, $k$ is the number of attributes, $m$ is the index of the attribute being edited, $p_{j, c}^{\prime}$ is the softmax output of predictor $P^{\prime}, $ $y_{j, c}$ is the binary indicator with respect to the target class and it is obtained by feeding the original image to the attribute predictor.

\subsection{Implementation Details on Comparison Methods}

\noindent\textbf{InterfaceGAN.}  InterfaceGAN \cite{interfacegan1} is a latent space based method. The continuous editing is achieved by moving the latent code along a straight line, \ie, adding the a same vector to the original latent code.
The direction used for changing the attribute degree is obtained by computing the normal vector of the binary classification SVM boundary. This direction is fixed throughout the editing.
We first train binary attribute predictors to classify the generated images. Then the corresponding latent codes are used to train the binary SVM.

\noindent\textbf{Multiclass SVM.} We further propose an extended version of InterfaceGAN as one of the baseline methods, named Multiclass SVM. Instead of the binary classification SVM, we train multiple SVM boundaries for fine-grained labels. More specifically, for each pair of neighbouring classes, a classification SVM would be trained. Thus, for one attribute, there are five SVM boundaries in total. During the editing, directions will be switched according to current states.
The attribute predictor used for the classification of generated images is the same as the one we use for predictor loss.

\noindent\textbf{Enjoy Your Editing.} Enjoy your editing \cite{enjoy_your_editing} learns a mapping network to generate identity-specific directions for each initial latent codes.
The identity-specific directions keep same during editing for one image. We reimplement the method, train the mapping network with the original design and same hyper-parameters are adopted. 
To achieve more attribute degrees, we use larger step-sizes than the original setting, \ie $\varepsilon > 1.0$. 

\section{More Qualitative Results}
\label{qualitative}
\begin{figure*}
   \begin{center}
      \includegraphics[width=0.95\linewidth]{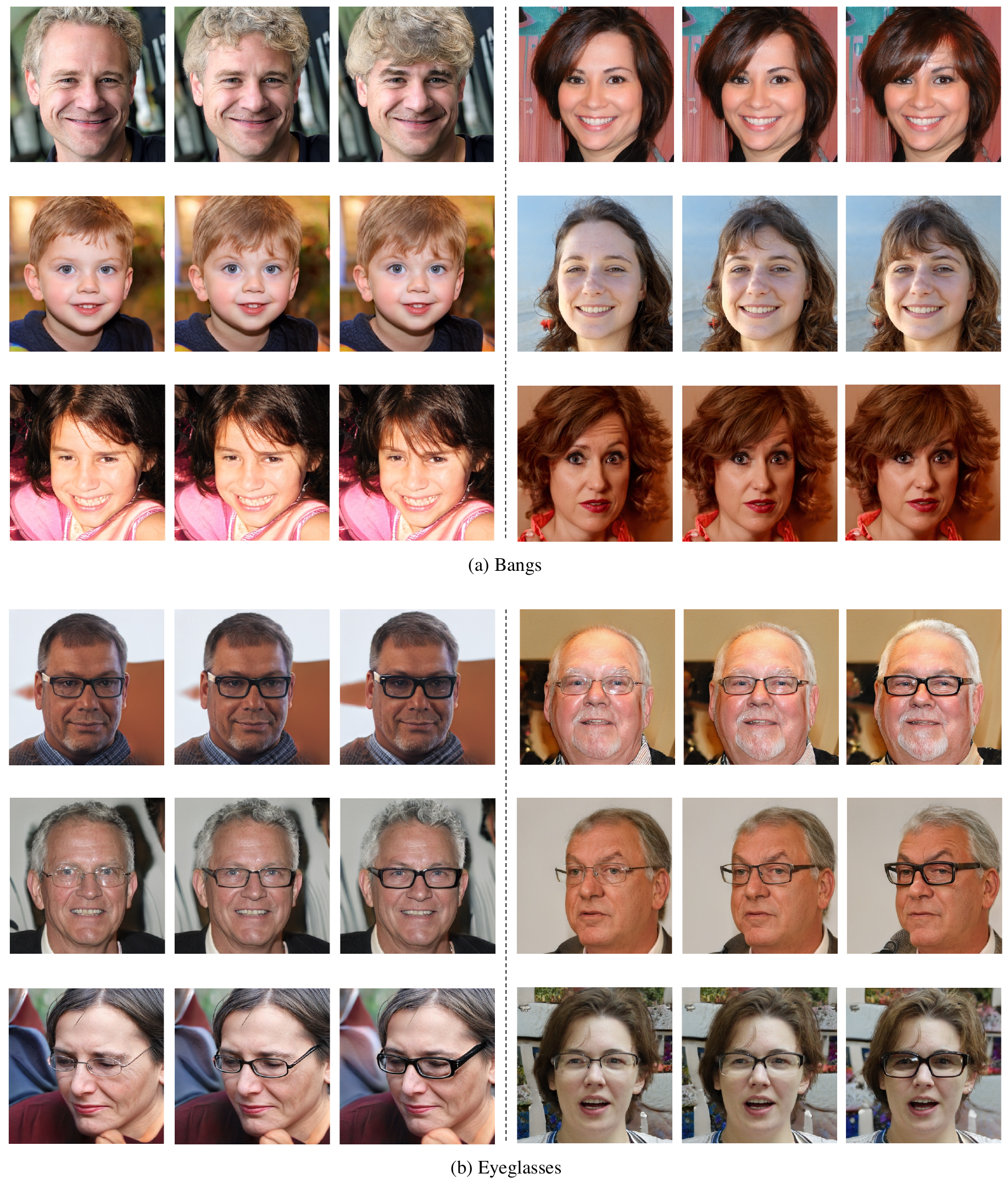}
   \end{center}
   \vspace{-0.5cm}
   \caption{\textbf{High-Resolution Image Editing.}}
   \label{HQ_results_1}
\end{figure*}

\begin{figure*}
   \begin{center}
      \includegraphics[width=0.95\linewidth]{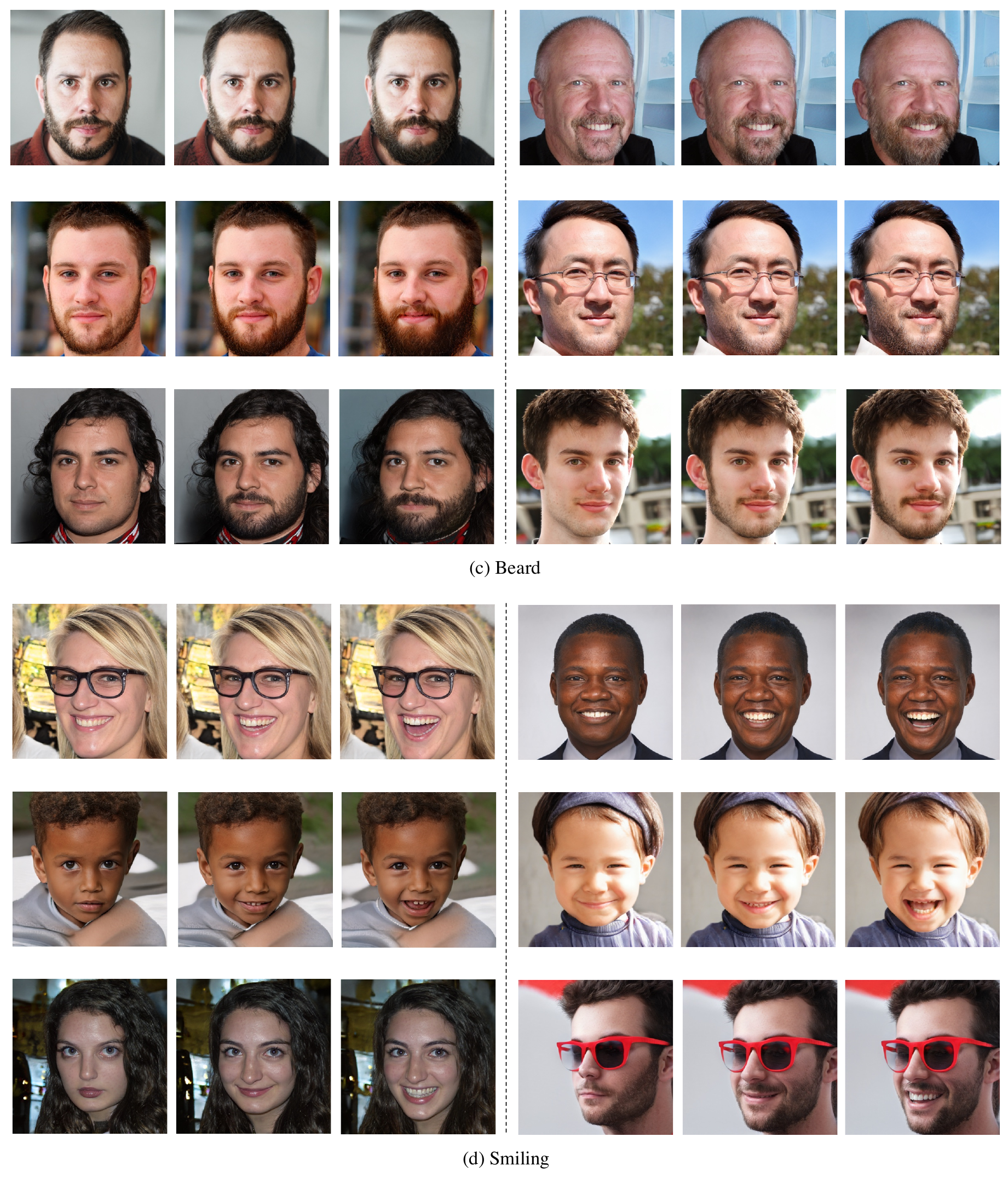}
   \end{center}
   \vspace{-0.5cm}
   \caption{\textbf{High-Resolution Image Editing.}}
   \label{HQ_results_2}

\end{figure*}

\begin{figure*}
  \begin{center}
      \includegraphics[width=0.95\linewidth]{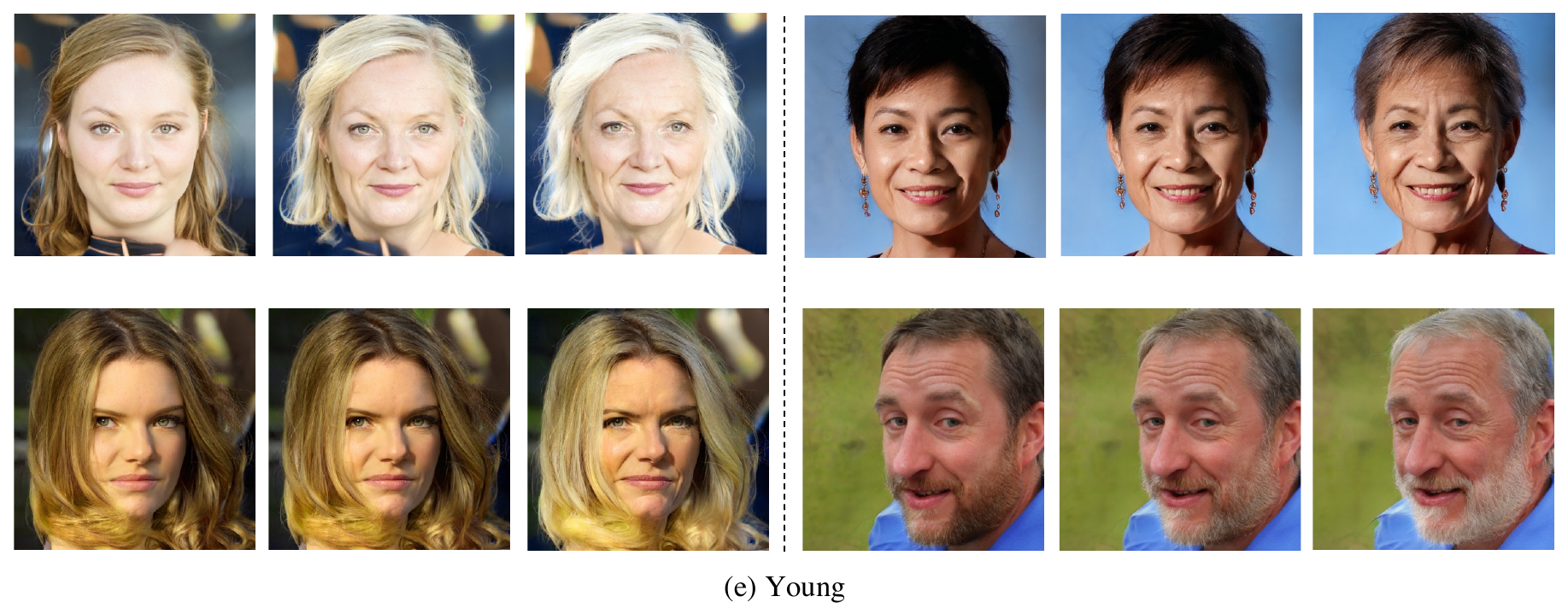}
  \end{center}
  \vspace{-0.5cm}
  \caption{\textbf{High-Resolution Image Editing.}}
  \label{HQ_results_3}
  \vspace{3cm}
  \begin{center}
      \includegraphics[width=0.95\linewidth]{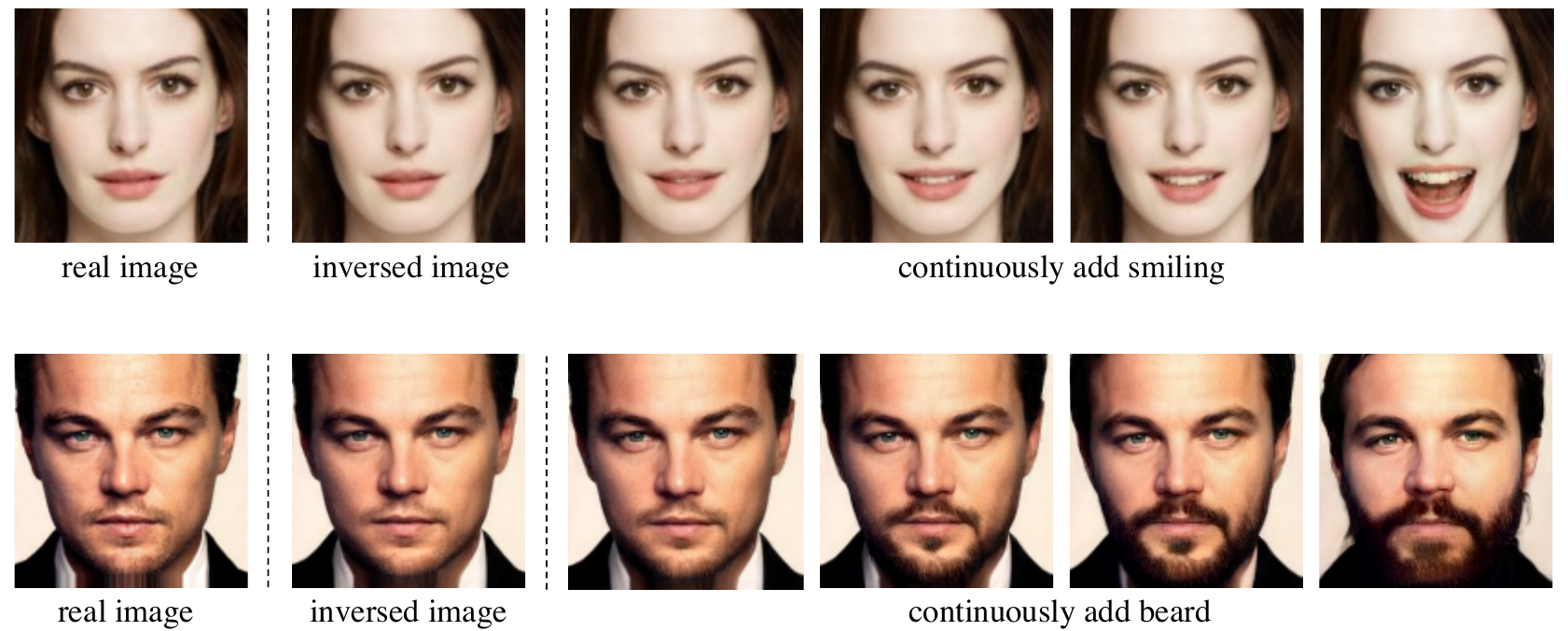}
   \end{center}
   \vspace{-0.5cm}
   \caption{\textbf{Real Image Editing.}}
   \label{real_img}
\end{figure*}

\begin{figure*}
  \begin{center}
      \includegraphics[width=1.0\linewidth]{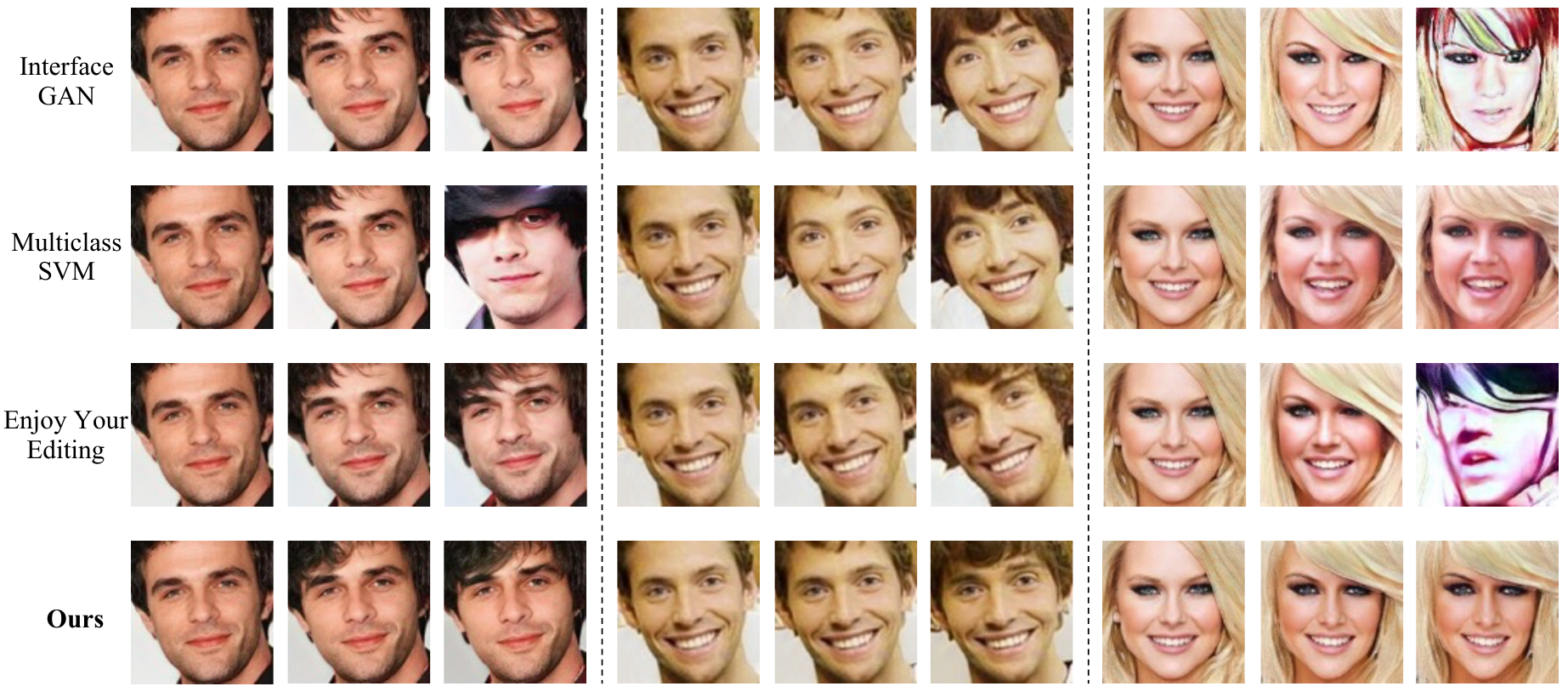}
  \end{center}
  \vspace{-0.5cm}
  \caption{\textbf{Qualitative Comparison on \textit{Bangs} Attribute.}}
  \label{bangs_qualitative}
%   \vspace{1cm}
  \begin{center}
      \includegraphics[width=1.0\linewidth]{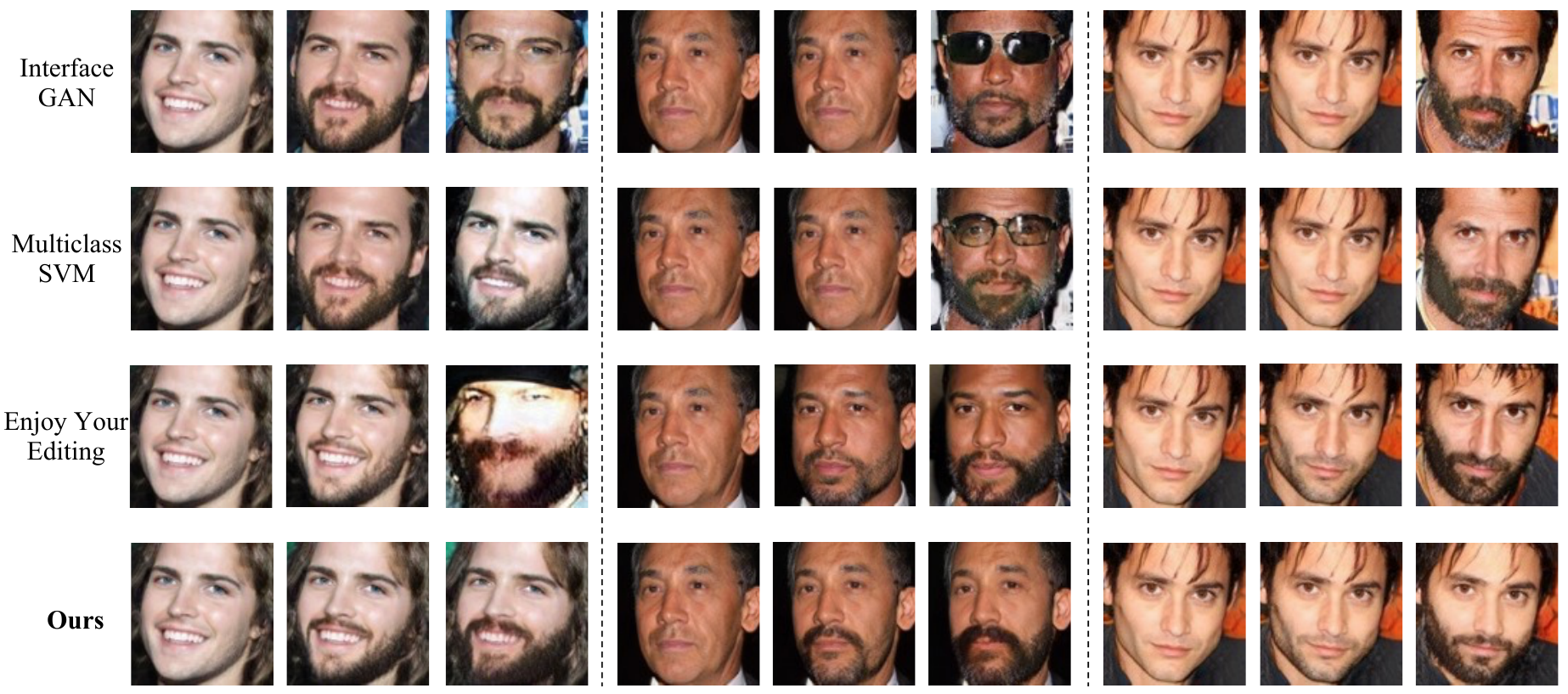}
  \end{center}
  \vspace{-0.5cm}
  \caption{\textbf{Qualitative Comparison on \textit{Beard} Attribute.}}
  \label{beard_qualitative}
\end{figure*}

\begin{figure*}
  \begin{center}
      \includegraphics[width=1.0\linewidth]{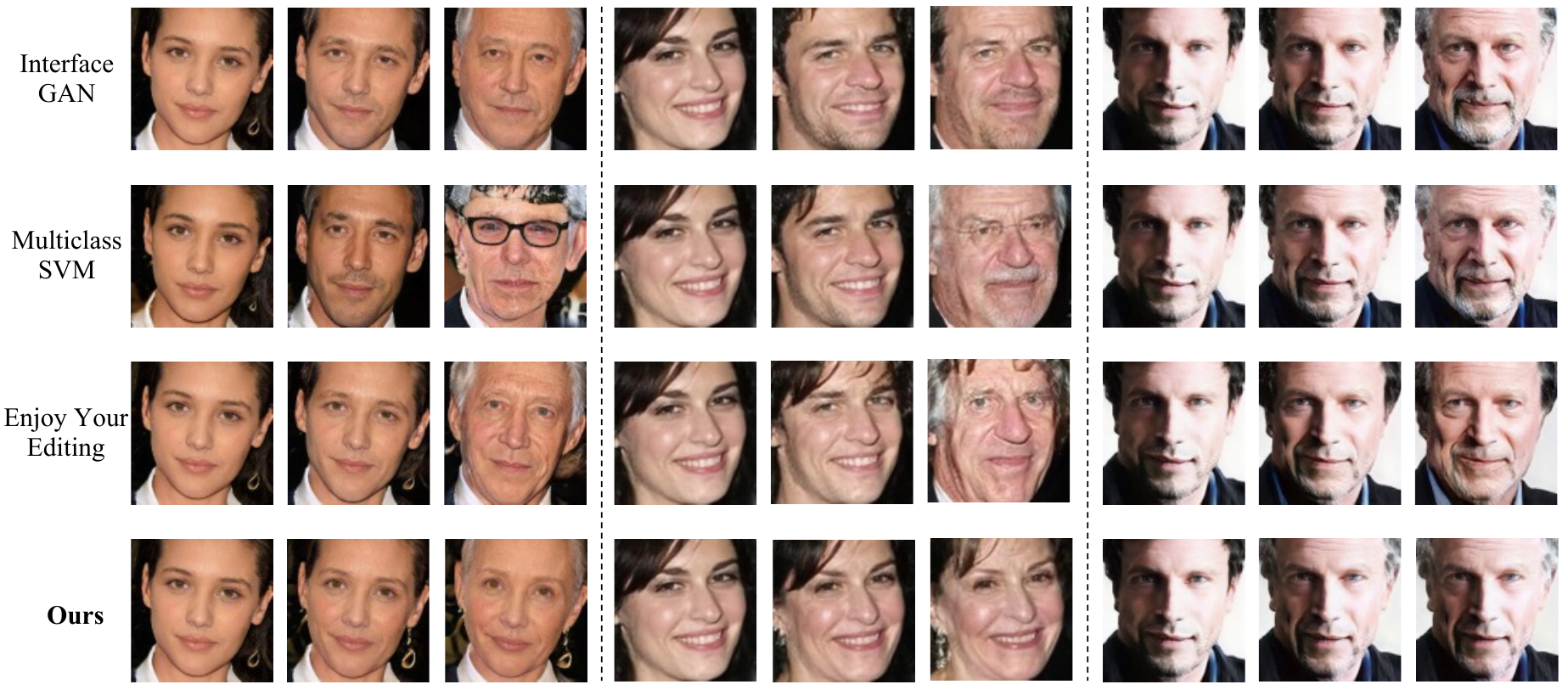}
  \end{center}
  \vspace{-0.5cm}
  \caption{\textbf{Qualitative Comparison on \textit{Young} Attribute.}}
  \label{age_qualitative}
\end{figure*}

In this section, we will provide more high-resolution image editing results in Fig. \ref{HQ_results_1}, \ref{HQ_results_2}, and Fig. \ref{HQ_results_3}.
We also provide more real image editing results in Fig. \ref{real_img} and more qualitative comparisons with baselines in Fig. \ref{bangs_qualitative} - Fig. \ref{age_qualitative}.

\section{Failure Cases Discussion}
\label{failur_discuss}
\begin{figure*}[h]
   \begin{center}
      \includegraphics[width=0.95\linewidth]{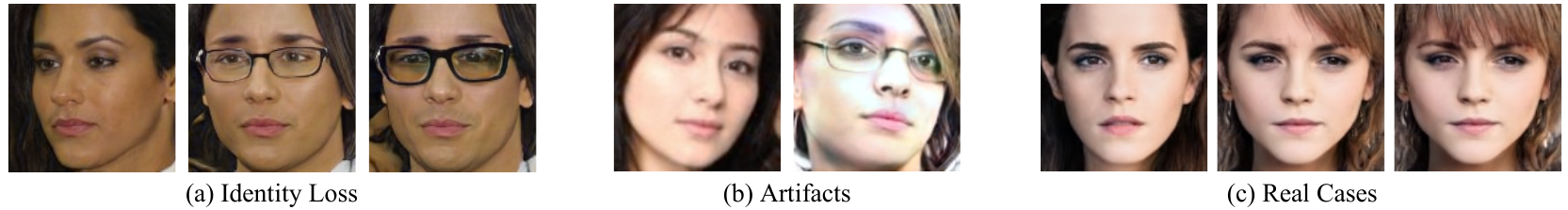}
   \end{center}
   \vspace{-0.6cm}
   \caption{\textbf{Failure Case Discussion.}}
   \label{failure_case}
\end{figure*}

Here, we take the eyeglasses attribute as an example to illustrate the failure case of synthetic image editing. As shown in Fig. \ref{failure_case} (a), 
identity loss could be observed in some cases, and this issue is severer on female images. 
The problem may attribute to the dataset bias and the mode collapse issue of the pretrained GAN. 
For example, the CelebA dataset \cite{celeba} has only a small number of females with eyeglasses. 
Thus, females with eyeglasses are only a minority in the image distribution of the pretrained GAN.
In this case, given a randomly sampled female without eyeglasses as the initial image, it is sometimes difficult to wear a pair of eyeglasses for her in a well-disentangled manner. 
Another issue is the artifacts problem shown in Fig. \ref{failure_case} (b). For some latent code, it is difficult to change the attribute from degree 0 to degree 1. After many latent code updating iterations, the latent code falls into the outlier region of the latent space so that the corresponding image would bear artifacts.
Our proposed semantic field may not perfectly model the non-linearity property for this attribute.

As for editing real images, it is more prone to change the identities. As shown in Fig. \ref{failure_case} (c), adding bangs would change the face shape. 
This is because that GAN-inversion, as an ill-posed problem, may introduce an additional gap between the inverted latent code and the original latent space.
This could potentially be addressed by adopting more advanced GAN-inversion techniques that better keep the latent codes within the latent domain.

\vspace{-0.3cm}

\begin{figure*}
  \begin{center}
      \includegraphics[width=1.0\linewidth]{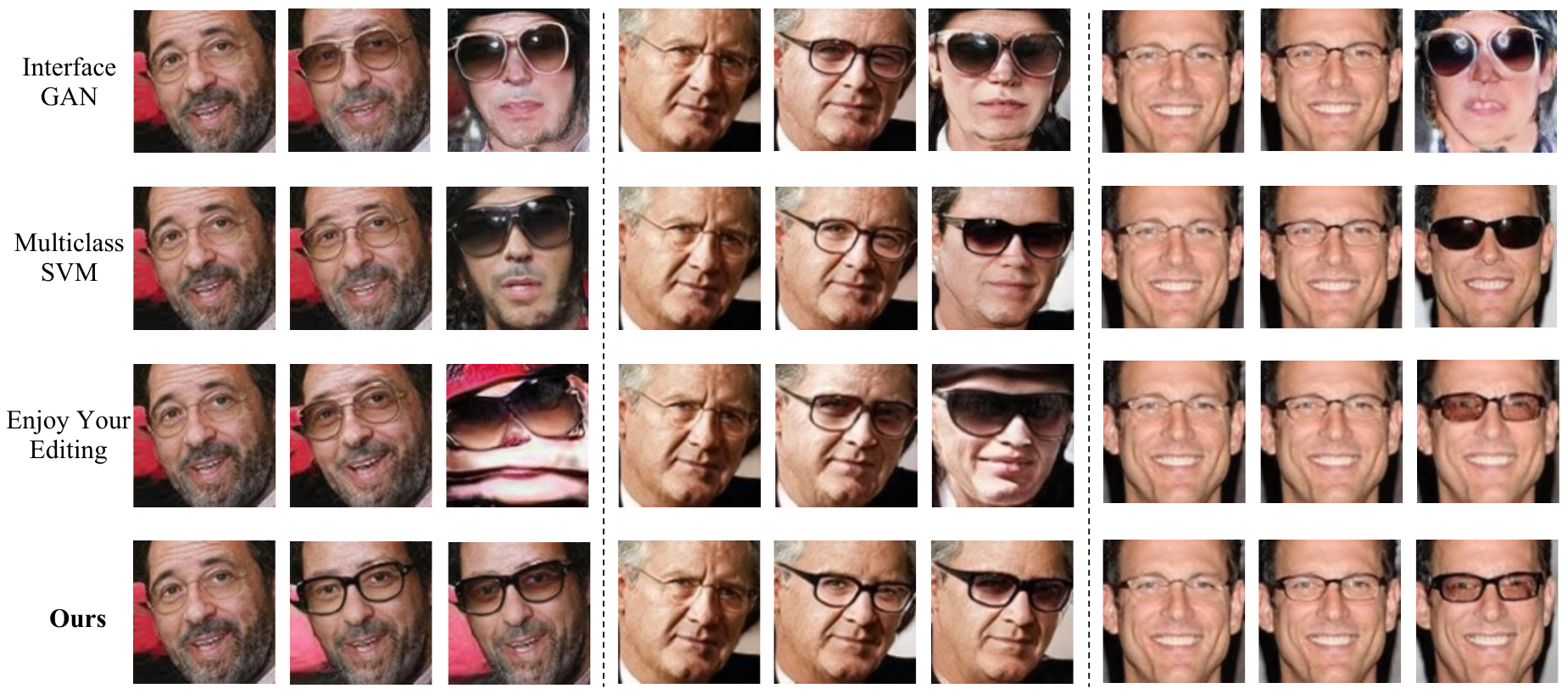}
  \end{center}
  \vspace{-0.5cm}
  \caption{\textbf{Qualitative Comparison on \textit{Eyeglasses} Attribute.}}
  \label{eyeglasses_qualitative}
  \vspace{1cm}
  \begin{center}
      \includegraphics[width=1.0\linewidth]{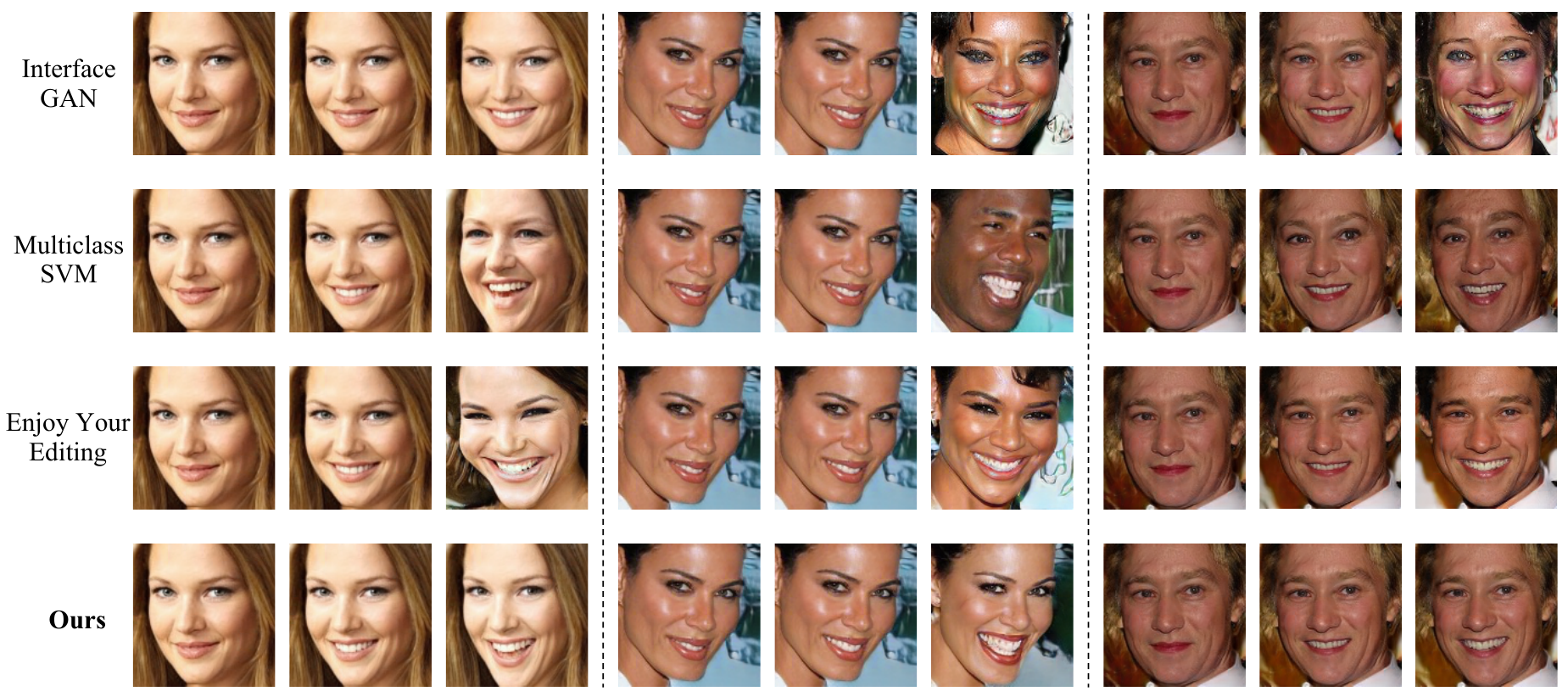}
  \end{center}
  \vspace{-0.5cm}
  \caption{\textbf{Qualitative Comparison on \textit{Smiling} Attribute.}}
  \label{eyeglasses_qualitative}
\end{figure*}

